\documentclass[10pt,twocolumn,letterpaper]{article}

\usepackage{cvpr}
\usepackage{times}
\usepackage{epsfig}
\usepackage{graphicx}
\usepackage{amsmath}
\usepackage{amssymb}

\usepackage{booktabs}
\usepackage{tablefootnote}
\usepackage{comment}
\usepackage{verbatim}
\usepackage{subcaption}
\usepackage{mathtools}
\usepackage{makecell}
\usepackage{multirow}
\usepackage{color,soul}
\usepackage{bbm}
\usepackage[dvipsnames]{xcolor}
\usepackage{array}

\newcommand{\head}[1]{\noindent\textbf{#1}}
\newcommand{\ourdataset}{ActivityNet-Entities\xspace}
\newcommand{\ourdatasetabbrev}{ANet-Entities\xspace}
\newcommand{\defeq}{\vcentcolon=}

\newcommand{\GVD}{GVD\xspace}
 \newcommand{\GID}{GVD\xspace}

\usepackage[pagebackref=true,breaklinks=true,letterpaper=true,colorlinks,bookmarks=false]{hyperref}

\cvprfinalcopy 

\def\cvprPaperID{12} 

\hypersetup{
    colorlinks=true,
    linkcolor=BrickRed,
    filecolor=magenta,      
    urlcolor=BrickRed,
}

\ifcvprfinal\fi
\begin{document}

\title{Grounded Video Description}

\author{
Luowei Zhou$^{1,2}$, Yannis Kalantidis$^1$, Xinlei Chen$^1$, Jason J. Corso$^2$, Marcus Rohrbach$^1$  \\
$^1$ Facebook AI, $^2$ University of Michigan \\
\href{https://github.com/facebookresearch/grounded-video-description}{github.com/facebookresearch/grounded-video-description}
}

\setlength{\abovedisplayskip}{3pt}
\setlength{\belowdisplayskip}{3pt}

\maketitle

\begin{abstract}
Video description is one of the most challenging problems in vision and language understanding due to the large variability both on the video and language side. Models, hence, typically shortcut the difficulty in recognition and generate plausible sentences that are based on priors but are not necessarily grounded in the video.
In this work, we explicitly link the sentence to the evidence in the video
by annotating each noun phrase in a sentence with the corresponding bounding box in one of the frames of a video. Our dataset, \ourdataset, augments the challenging ActivityNet Captions dataset with 158k bounding box annotations, each grounding a noun phrase. This allows training video description models with this data, and importantly, evaluate how grounded or ``true'' such model are to the video they describe. To generate grounded captions, we propose a novel  
video description model which is able to exploit these bounding box annotations. We demonstrate the effectiveness of our model on our dataset, but also show how it can be applied to image description on the Flickr30k Entities dataset. We achieve  state-of-the-art performance on video description, video paragraph description, and image description and demonstrate our generated sentences are better grounded in the video. 
\end{abstract}

\vspace{-0.25in}
\section{Introduction}
\label{sec:intro}

Image and video description models are frequently not well grounded \cite{liu2017attention} which can increase their bias \cite{hendricks18eccv} and lead to hallucination of objects \cite{rohrbach18emnlp}, \ie the  model mentions objects which are not in the image or video \eg because they might have appeared in similar contexts during training. This makes models less accountable and trustworthy, which is important if we hope such models will eventually assist people in need \cite{bigham10uist,rohrbach17ijcv}. Additionally, grounded models can help to explain the model's decisions to humans and allow humans to diagnose them \cite{park18cvpr}.
While researchers have started to discover and study these problems for image description \cite{liu2017attention,hendricks18eccv,rohrbach18emnlp,park18cvpr},\footnote{We use \emph{description} instead of \emph{captioning} as \emph{captioning} is often used to refer to transcribing the speech in the video, not \emph{describing} the content.} they are even more pronounced for video description due to the increased difficulty and diversity, both on the visual and the language side.

\begin{figure}[t]
\centering
\includegraphics[width=0.95\linewidth]{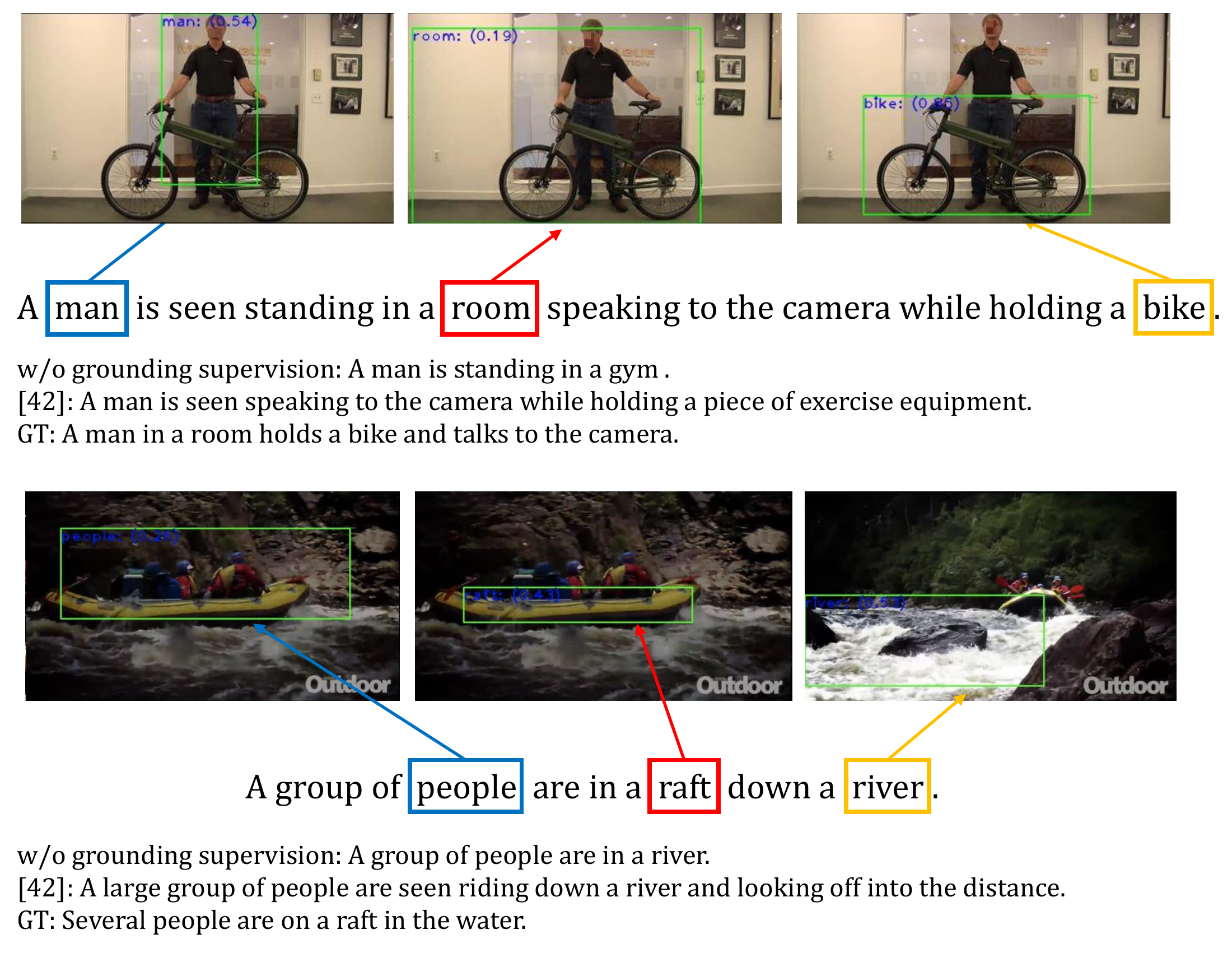}
\caption{
   Word-level grounded video descriptions generated by our model on two segments from our \ourdataset dataset. We also provide the descriptions generated by our model without explicit bounding box supervision, the descriptions generated by~\cite{zhou2018end} and the ground-truth descriptions (GT) for comparison. 
}
\label{fig:fig1}
\vspace{-10pt}
\end{figure}

Fig.~\ref{fig:fig1} illustrates this problem. A video description approach (without grounding supervision)
generated the sentence ``A man standing in a gym''
which correctly mentions ``a man'' but hallucinates ``gym'' which is not visible in the video. Although a man is in the video it is not clear if the model looked at the bounding box of the man to say this word~\cite{hendricks18eccv,rohrbach18emnlp}. For the sentence ``A man [...] is playing the  piano'' in Fig.~\ref{fig:dataset}, it is important to understand that which ``man'' in the image ``A man''
is referring to, to determine if a model is correctly grounded. Such understanding is crucial for many applications when trying to build accountable systems or  when generating the next sentence or responding to a follow up question of a blind person: \eg  answering  ``Is \emph{he} looking at me?'' requires an understanding which of the people in the image the model talked about.

The goal of our research is to  build such grounded systems. As one important step in this direction, we collect \ourdataset (short as \ourdatasetabbrev) which grounds or links noun phrases in sentences with bounding boxes in the video frames.
It is based on ActivityNet Captions~\cite{krishna2017dense}, one of the largest benchmarks in video description.
When annotating objects or noun phrases we specifically annotate the bounding box which corresponds to the instance referred to in the sentence rather than all instances of the same object category, \eg in Fig.~\ref{fig:dataset}, for the noun phrase ``the man'' in the video description, we only annotate the sitting man and not the standing man or the woman, although they are all from the object category ``person''.
We note that annotations are sparse, in the sense that we only annotate a single frame of the video for each noun phrase.
\ourdatasetabbrev has a total number of 51.8k annotated video segments/sentences with 157.8k labeled bounding boxes, more details can be found in Sec.~\ref{sec:anet-bb}.

Our new dataset allows us to introduce a novel grounding-based video description model that learns to jointly generate words and refine the grounding of the objects generated in the description. We explore how this explicit supervision can benefit the description generation compared to unsupervised methods that might also utilize region features but do not penalize grounding. 

Our contributions are three-fold. First, we collect our large-scale  \ourdataset dataset, which grounds video descriptions to bounding boxes on the level of noun phrases. Our dataset allows both, \emph{teaching} models to explicitly rely on the corresponding evidence in the video frame when generating words and \emph{evaluating} how well models are doing in grounding individual words or phrases they generated.
Second, we propose a grounded video description framework which is able to learn from the bounding box supervision in \ourdataset and we demonstrate its superiority over baselines and prior work in generating grounded video descriptions.
Third, we show the applicability of the proposed model to image captioning, again showing improvements in the generated captions and the quality of grounding on the Flickr30k Entities~\cite{plummer2015flickr30k} dataset.

\vspace{-0.1in}
\section{Related Work}
\label{sec:related}

\head{Video \& Image  Description.} Early work on automatic caption generation mainly includes template-based approaches~\cite{das2013thousand, kulkarni2013babytalk, mitchell2012midge}, where predefined templates with slots are first generated and then filled in with detected visual evidences. Although these works tend to lead to well-grounded methods, they are restricted by their template-based nature. More recently, neural network and attention-based methods have started to dominate major captioning benchmarks. Visual attention usually comes in the form of temporal attention~\cite{yao2015describing} (or spatial-attention~\cite{xu2015show} in the image domain), semantic attention~\cite{li2018jointly, yao2017boosting, you2016image, zhou2017watch} or both~\cite{pan2017video}. The recent unprecedented success in object detection~\cite{ren2015faster, he2017mask} has regained the community's interests on detecting fine-grained visual clues while incorporating them into end-to-end networks~\cite{ma2018attend, rohrbach2017generating, anderson2018bottom, lu2018neural}.
Description methods which are based on object detectors~\cite{ma2018attend, zanfir16accv, anderson2018bottom, lu2018neural,das2013thousand,kulkarni2013babytalk} tackle the captioning problem in two stages. They first use off-the-shelf or fine-tuned object detectors to propose object proposals/detections as for the visual recognition heavy-lifting. Then, in the second stage, they either attend to the object regions dynamically~\cite{ma2018attend, zanfir16accv, anderson2018bottom} or classify the regions into labels and fill into pre-defined/generated sentence templates~\cite{lu2018neural,das2013thousand,kulkarni2013babytalk}. 
However, directly generating proposals from off-the-shelf detectors causes the proposals to bias towards classes in the source dataset (\ie for object detection) v.s. contents in the target dataset (\ie for description). One solution is to fine-tune the detector specifically for a dataset~\cite{lu2018neural} but this requires exhaustive object annotations that are difficult to obtain, especially for videos.
Instead of fine-tuning a general detector, we transfer the object classification knowledge from off-the-shelf object detectors to our model and then fine-tune this representation as part of our generation model with sparse box annotations.
With a focus on co-reference resolution and identifying people, \cite{rohrbach2017generating} proposes a framework that can refer to particular character instances and do visual co-reference resolution between video clips. However, their method is restricted to identifying human characters whereas we study more general the grounding of objects.

\head{Attention Supervision.} As fine-grained grounding becomes a potential incentive for next-generation vision-language systems, to what degree it can benefit remains an open question. On one hand, for 
VQA~\cite{das2017human,zhang2019interpretable} 
the authors point out that the attention model does not attend to same regions as humans and adding attention supervision barely helps the performance. On the other hand, adding supervision to feature map attention~\cite{liu2017attention, yu2017supervising} was found 
to be beneficial. We noticed in our preliminary experiments that directly guiding the region attention with supervision~\cite{lu2018neural} does not necessary lead to improvements in automatic sentence metrics. 
We hypothesize that this might be due to the lack of object context information and we thus introduce a self-attention~\cite{vaswani2017attention} based context encoding 
in our attention model, 
which allows information passing across 
all regions in the sampled video frames.

\vspace{-0.1in}
\section{\ourdataset Dataset}
\label{sec:anet-bb}

\begin{figure}[t]
\centering
\includegraphics[width=1\linewidth]{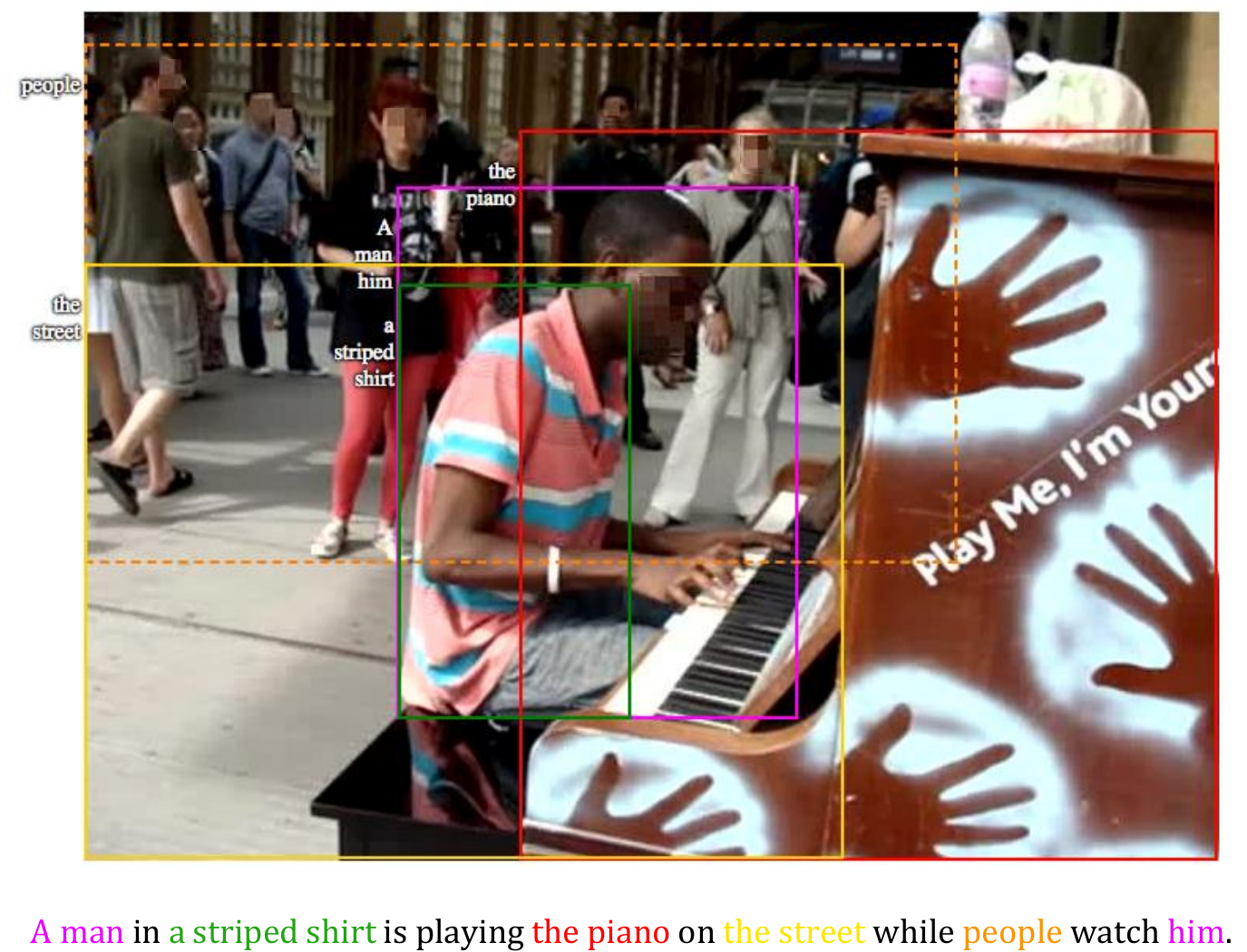}
\caption{An annotated example from our dataset. The dashed box (``people'') indicates a group of objects.}
\label{fig:dataset}
\vspace{-10pt}
\end{figure}

In order to train and test models capable of explicit grounding-based video description, one requires both language and grounding supervision. 
Although Flickr30k Entities~\cite{plummer2015flickr30k} contains such annotations for images, no large-scale description datasets with object localization annotation exists for videos. The large-scale ActivityNet Captions dataset~\cite{krishna2017dense} contains dense language annotations for about 20k videos from ActivityNet~\cite{caba2015activitynet} but lacks grounding annotations. 
Leveraging the language annotations from the ActivityNet Captions dataset~\cite{krishna2017dense}, we 
collected entity-level bounding box annotations and created the \ourdataset (\ourdatasetabbrev) dataset\footnote{\ourdataset is released at \url{https://github.com/facebookresearch/ActivityNet-Entities}.}, a rich dataset that can be used for video description with explicit grounding.
With 15k videos and more than 158k annotated bounding boxes, \ourdataset is the largest annotated dataset of its kind to the best of our knowledge.

When it comes to videos, region-level annotations come with a number of unique challenges. A video contains more information than can fit in a single frame, and video descriptions reflect that. They may reference objects that appear in a disjoint set of frames, as well as multiple persons and motions. To be more precise and produce finer-grained annotations, we annotate \emph{noun phrases} (NP) (defined below) rather than simple object labels. Moreover, one would ideally have dense region annotations at every frame, but the annotation cost in this case would be prohibitive for even small datasets. Therefore in practice, video datasets are typically sparsely annotated at the region level~\cite{gu2017ava}. Favouring scale over density, we choose to annotate segments as sparsely as possible and annotate every noun phrase only in one frame inside each segment.

\head{Noun Phrases}. Following~\cite{plummer2015flickr30k}, we define noun phrases as short, non-recursive phrases that refer to a specific region in the image, able to be enclosed within a bounding box. They can contain a single instance or a group of instances and may include adjectives, determiners, pronouns or prepositions. For granularity, we further encourage the annotators to split complex NPs into their simplest form (\eg ``the man in a white shirt with a heart'' can be split into three NPs: ``the man'', ``a white shirt'', and ``a heart'').

\subsection{Annotation Process} 

We uniformly sampled 10 frames from each video segment and presented them to the annotators together with the corresponding sentence. We asked the annotators to identify all concrete NPs from the sentence describing the video segment and then draw bounding boxes around them in \textit{one} frame of the video where the target NPs can be clearly observed. Further instructions were provided including guidelines for resolving co-references within a sentence, \ie boxes may correspond to multiple NPs in the sentence (\eg, a single box could refer to both ``the man'' and ``him'') or when to use \emph{multi-instance boxes} (\eg ``crowd'', ``a group of people'' or ``seven cats''). An annotated example is shown in Fig.~\ref{fig:dataset}. It is noteworthy that 10\% of the final annotations refer to multi-instance boxes.
We trained annotators, and deployed a rigid quality control by daily inspection and feedback. All annotations were verified in a second round. The full list of instructions provided to the annotators, validation process, as well as screen-shots of the annotation interface can be found in the Appendix. 

\subsection{Dataset Statistics and Analysis}
\label{sec:datasetstats}

\begin{table}[t]
\centering
\resizebox{1.0\columnwidth}{!}{
    \begin{tabular}{l|rrrrr}
        \toprule
        Dataset & Domain & \# Vid/Img & \# Sent  & \# Obj & \# BBoxes\\
        \midrule
        Flickr30k Entities~\cite{plummer2015flickr30k} & Image & 32k & 160k & 480 & 276k \\ 
        \midrule
        MPII-MD~\cite{rohrbach2017generating} & Video & $\ll$1k & $\ll$1k & 4 &  2.6k \\
        YouCook2~\cite{zhou2018weakly} & Video & 2k & 15k & 67 & 135k \\
        ActivityNet Humans~\cite{yamaguchi2017spatio} & Video & 5.3k & 30k & 1 & 63k \\
        \textbf{\ourdataset (ours)} & \textbf{Video} & \textbf{15k} &  \textbf{52k} & \textbf{432} & \textbf{158k} \\ 
        \quad --train & & 10k & 35k & 432 & 105k \\
        \quad --val &  & 2.5k & 8.6k & 427 & 26.5k \\
        \quad --test &  & 2.5k & 8.5k & 421 & 26.1k \\
        \bottomrule
    \end{tabular}
}
\vspace{-5pt}
  \caption{Comparison of video description datasets with  noun phrase or word-level grounding annotations. 
  Our \ourdataset and ActivityNet Humans~\cite{yamaguchi2017spatio} dataset are both based on ActivityNet~\cite{caba2015activitynet}, but ActivityNet Humans provides boxes only for person on a small subset of videos. YouCook2 is restricted to cooking and only has box annotations for the val and the test splits.}\label{tab:dataset}
\vspace{-10pt}
\end{table}

As the test set annotations for the ActivityNet Captions dataset are not public, we only annotate the segments in the training (train) and validation (val) splits. This brings the total number of annotated videos in \ourdataset to 14,281. In terms of segments, we ended up with about 52k video segments with at least one NP annotation and 158k NP bounding boxes in total.

Respecting the original protocol, we keep as our training set the corresponding split from the ActivityNet Captions dataset. We further randomly \& evenly split the original val set into our val set and our test set. We use all available bounding boxes for training our models, \ie, including multi-instance boxes. 
Complete stats and comparisons with other related datasets can be found in Tab.~\ref{tab:dataset}.

\begin{figure*}[t!]
\centering
   \includegraphics[width=.9\linewidth]{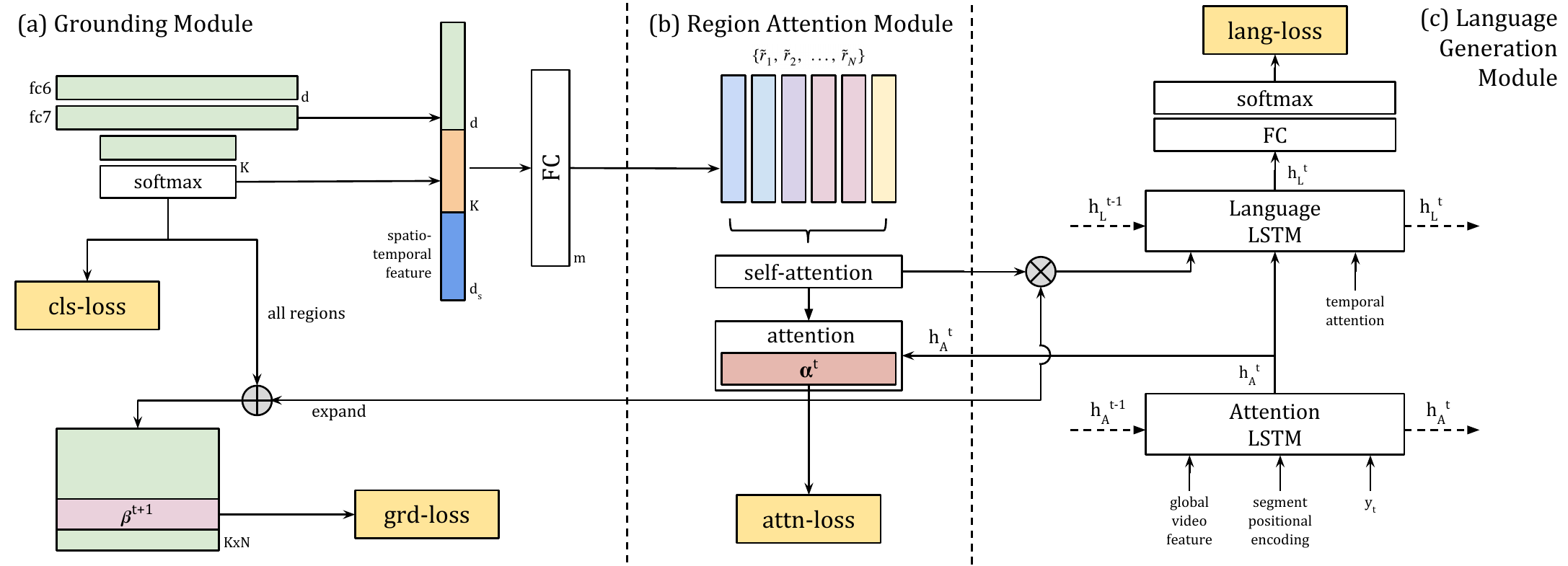}
   \caption{The proposed framework consists of three parts: the grounding module (a), the region attention module (b) and the language generation module (c). Region proposals are first represented with grounding-aware region encodings. The language model then dynamically attends on the region encodings to generate each word. Losses are imposed on the attention weights (attn-loss), grounding weights (grd-loss), and the region classification probabilities (cls-loss). For clarity, the details of the temporal attention are omitted.}
   \vspace{-10pt}
\label{fig:model}
\end{figure*}

\head{From Noun Phrases to Objects Labels}. Although we chose to annotate noun phrases, in this work, we model sentence generation as a word-level task. We follow the convention in~\cite{lu2018neural} to determine the list of object classes and convert the NP label for box to a single-word object label. First, we select all nouns and pronouns from the NP annotations using the Stanford Parser~\cite{manning2014stanford}. The frequency of these words in the train and val splits are computed and a threshold determines whether each word is an object class. For \ourdatasetabbrev, we set the frequency threshold to be 50 which produces 432 object classes.

\vspace{-0.1in}
\section{Description with Grounding Supervision}\label{sec:method}

In this section we describe the proposed grounded video description framework (see Fig.~\ref{fig:model}). The framework consists of three modules: grounding, region attention and language generation. The grounding module detects visual clues from the video, the region attention dynamically attends on the visual clues to form a high-level impression of the visual content and feeds it to the language generation module for decoding. We illustrate three options for incorporating the object-level supervision: region classification, object grounding (localization), and supervised attention.

\subsection{Overview}

We formulate the problem as a joint optimization over the language and grounding tasks. The overall loss function consists of four parts:
\begin{equation}
L=L_{sent}+\lambda_{\alpha} L_{attn}+\lambda_{c} L_{cls}+\lambda_{
\beta} L_{grd},
\end{equation}
where $L_{sent}$ denotes the teacher-forcing language generation cross-entropy loss, commonly used for language generation tasks (details in Sec.~\ref{sec:lang_gen_module}). $L_{attn}$ corresponds to the cross entropy region attention loss 
which is presented in Sec.~\ref{sec:reg_att_module}.
$L_{cls}$ and $L_{grd}$ are cross-entropy losses that correspond to the grounding module for region classification 
and supervised object grounding (localization),
 respectively (Sec.~\ref{sec:grounding_module}).
The three grounding-related losses are weighted by coefficients $\lambda_\alpha$, $\lambda_c$, and $\lambda_\beta$ which we selected on the dataset validation split.

We denote the input video (segment) as $V$ and the target/generated sentence description (words) as $S$. We uniformly sample $F$ frames from each video as $\{v_1, v_2, \ldots, v_F\}$ and define $N_f$ object regions on sampled frame $f$. Hence, we can assemble a set of regions $R=[R_1, \ldots, R_F]=[r_1, r_2, \ldots, r_{N}] \in \mathbb{R}^{d\times N}$ to represent the video, where $N=\sum_{f=1}^FN_f$ is the total number of regions. We overload the notation here and use $r_i$ ($i\in\{1,2,\ldots,N\}$) to also represent region feature embeddings, as indicated by fc6 in Fig.~\ref{fig:model}. We represent words in $S$ with one-hot vectors which are further encoded to word embeddings $y_t \in \mathbb{R}^{e}$ where $t\in \{ 1,2, \ldots, T\}$, where $T$ indicates the sentence length and $e$ is the embedding size.

\subsection{Language Generation Module}
\label{sec:lang_gen_module}
For language generation, we adapt the language model from~\cite{lu2018neural} for video inputs, \ie extend it to incorporate temporal information. The model consists of two LSTMs: the first one for encoding the global video feature and the word embedding $y_t$
into the hidden state $h_A^t\in \mathbb{R}^m$ where $m$ is the dimension and the second one for language generation (see Fig.~\ref{fig:model}c). The language model dynamically attends on videos frames or regions for visual clues to generate words. We refer to the attention on video frames as temporal attention and the one on regions as region attention.

The temporal attention takes in a sequence of frame-wise feature vectors and determines by the hidden state how significant each frame should contribute to generate a description word. We deploy a similar module as in~\cite{zhou2018end}, except that we replace the self-attention context encoder with Bi-directional GRU (Bi-GRU) which yields superior results. We train with
cross-entropy loss $L_{sent}$. 

\subsection{Region Attention Module}
\label{sec:reg_att_module}

Unlike temporal attention that works on a frame level, the region attention~\cite{anderson2018bottom, lu2018neural} focuses on more fine-grained details in the video, \ie, object regions~\cite{ren2015faster}. We denote the 
region encoding  as $\Tilde{R}=[\Tilde{r}_1, \Tilde{r}_2, \ldots, \Tilde{r}_N]$, more details are defined later in Eq.~\ref{eq:region_enc}. At time $t$ of the caption generation, the attention weight over region $i$ is formulated as:
\begin{equation}\label{eq:alpha}
\alpha_i^t=w_\alpha^\top\text{tanh}(W_r \Tilde{r}_i+W_hh_A^t), \;\;\;\;
\alpha^t \defeq \text{Softmax}(\alpha^t),
\end{equation}
\noindent where $W_r\in \mathbb{R}^{m\times d}$, $W_h\in \mathbb{R}^{m\times m}$, $w_\alpha\in \mathbb{R}^{m}$, and $\alpha^t=[\alpha_1^t,\alpha_2^t,\ldots,\alpha_N^t]$. The region attention encoding is then $\tilde{R} \alpha^t$ and along with the temporal attention encoding, fed into the language LSTM. 

\head{Supervised Attention.} We want to encourage the language model to attend on the correct region when generating a visually-groundable word. As this effectively assists the language model in learning to attend to the correct region, we call this \textit{attention supervision}. Denote the indicators of positive/negative regions as $\gamma^t=[\gamma_1^t, \gamma_2^t, \ldots, \gamma_N^t]$, where $\gamma_i^t=1$ when the region $r_i$ has over 0.5 IoU with the GT box $r_{GT}$ and otherwise 0. We regress $\alpha^t$ to $\gamma^t$ and hence the attention loss for object word $s_{t}$ can be defined as:
\begin{equation}\label{eq:attn}
    L_{attn}=-\sum_{i=1}^N \gamma_i^t \log \alpha_i^t.
\end{equation}

\subsection{Grounding Module}
\label{sec:grounding_module}

Assume we have a set of visually-groundable object class labels $\{c_1, c_2,\ldots, c_\mathcal{K}\}$, short as object classes, where $\mathcal{K}$ is the total number of classes. Given a set of object regions from all sampled frames, the grounding module estimates the class probability distribution for each region.

We define a set of object classifiers as $W_c=[w_1, w_2, \ldots, w_\mathcal{K}]\in \mathbb{R}^{d\times \mathcal{K}}$ and the learnable scalar biases as $B=[b_1, b_2, \ldots, b_\mathcal{K}]$. So, a naive way to estimate the class probabilities for all regions (embeddings) $R=[r_1, r_2, \ldots, r_N]$ is through dot-product:
\begin{equation}\label{eq:ms}
M_s(R)=\text{Softmax}(W_c^\top R+B\mathbbm{1}^\top),
\end{equation}
where $\mathbbm{1}$ is a vector with all ones, $W_c^\top R$ is followed by a ReLU and a Dropout layer, and $M_s$ is the \textit{region-class similarity matrix} as it captures the similarity between regions and object classes. For clarity, we omit the ReLU and Dropout layer after the linear embedding layer throughout Sec.~\ref{sec:method} unless otherwise specified. The Softmax operator is applied along the object class dimension of $M_s$ to ensure the class probabilities for each region sum up to 1.

We transfer detection knowledge from an off-the-shelf detector that is pre-trained on a general source dataset, \ie, Visual Genome (VG)~\cite{krishna2017visual}, to our object classifiers. We find the nearest neighbor for each of the $\mathcal{K}$ object classes from the VG object classes according to their distances in the embedding space (glove vectors~\cite{pennington2014glove}). We then initialize $W_c$ and $B$ with the corresponding classifier, \ie, the weights and biases, from the last linear layer of the detector.

On the other hand, we represent the spatial and temporal configuration of the region as a 5-D tuple, including 4 values for the normalized spatial location and 1 value for the normalized frame index. Then, the 5-D feature is projected to a $d_s=300$-D location embedding for all the regions $M_l\in \mathbb{R}^{300 \times N}$. Finally, we concatenate all three components: i) region feature, ii) region-class similarity matrix, and iii) location embedding together and project into a lower dimension space (m-D):
\begin{equation}\label{eq:region_enc}
   \Tilde{R} = W_g [ \; R \; | \; M_s(R) \; | \; M_l \; ],
\end{equation}
\noindent where $[\cdot | \cdot]$ indicates a row-wise concatenation and $W_g\in \mathbb{R}^{m\times(d+K+d_s)}$ are the embedding weights. We name $\Tilde{R}$ the \textit{grounding-aware region encoding}, corresponding to the right portion of Fig.~\ref{fig:model}a. To further model the relations between regions, we deploy a self-attention layer over $\Tilde{R}$~\cite{vaswani2017attention,zhou2018end}.
The final region encoding is fed into the region attention module (see Fig.~\ref{fig:model}b).

So far the object classifier discriminates classes without the prior knowledge about the semantic context, \ie, the information the language model has captured. To incorporate semantics, we condition the class probabilities on the sentence encoding from the Attention LSTM. 
A memory-efficient approach is treating attention weights $\alpha^t$ as this semantic prior, as formulated below:
\begin{equation}\label{eq:mst}
M_s^t(R,\alpha^t)=\text{Softmax}(W_c^\top R+B\mathbbm{1}^\top+\mathbbm{1}{\alpha^t}^\top),
\end{equation}
\noindent where the region attention weights $\alpha^t$ are determined by Eq.~\ref{eq:alpha}. Note that here the Softmax operator is applied row-wise to ensure the probabilities on regions sum up to 1. To learn a reasonable object classifier, we can  deploy a region classification task on $M_s(R)$ or a sentence-conditioned grounding task on $M_s^t(R, \alpha^t)$, with the word-level grounding annotations from Sec.~\ref{sec:anet-bb}. Next, we describe them both.

\head{Region Classification.} We first define a positive region as a region that has  over 0.5 intersection over union (IoU) with an arbitrary ground-truth (GT) box. If a region matches to multiple GT boxes, the one with the largest IoU is the final matched GT box. Then we classify the positive region, say region $i$ to the same class label as in the GT box, say class $c_j$. The normalized class probability distribution is hence $M_s[:,i]$ and the cross-entropy loss on class $c_j$ is
\begin{equation}\label{eq:cls}
    L_{cls}= - \log M_s[j,i].
\end{equation}
\noindent The final $L_{cls}$ is the average of losses on all positive regions.

\head{Object Grounding.} Given a visually-groundable word $s_{t+1}$ at time step $t+1$ and the encoding of all the previous words, we aim to localize $s_{t+1}$ in the video as one or a few of the region proposals. Supposing $s_{t+1}$ corresponds to class $c_j$, we regress the confidence score of regions $M_s^t[j,:]=\beta^{t+1}=[\beta_1^{t+1}, \beta_2^{t+1},\ldots,\beta_N^{t+1}]$ to indicators $\gamma^t$ as defined in Sec.~\ref{sec:reg_att_module}. The grounding loss for word $s_{t+1}$ is defined as:
\begin{equation}\label{eq:grd}
    L_{grd}=-\sum_{i=1}^N \gamma_i^t \log \beta_i^{t+1}.
\end{equation}

Note that the final loss on $L_{attn}$ or $L_{grd}$ is the average of losses on all visually-groundable words.
The difference between the attention supervision and the grounding supervision is that, in the latter task, the target object $c_j$ is known beforehand, while the attention module is not aware of which object to seek in the scene.

\vspace{-0.1in}
\section{Experiments}\label{sec:exp}
\head{Datasets.}
We conduct most experiments and ablation studies on the newly-collected \ourdataset dataset on video description given the set of temporal segments (\ie using the ground-truth events from \cite{krishna2017dense}) and video paragraph description~\cite{xiong2018move}.
We also demonstrate our framework can easily be applied to image description and evaluate it on the Flickr30k Entities dataset~\cite{plummer2015flickr30k}. Note that we did not apply our method to COCO captioning as there is no exact match between words in COCO captions and object annotations in COCO (limited to only 80).
We use the same process described in Sec.~\ref{sec:datasetstats} to convert NPs to object labels. Since Flickr30k Entities contains more captions, labels that occur at least 100 times are taken as object labels, resulting in 480 object classes~\cite{lu2018neural}.

\head{Pre-processing.} 
For \ourdatasetabbrev, we truncate captions longer than 20 words and build a vocabulary on words with at least 3 occurrences. For Flickr30k Entities, since the captions are generally shorter and it is a larger corpus, we truncate captions longer than 16 words and build a vocabulary based on words that occur at least 5 times.

\subsection{Compared Methods and Metrics}
\label{sec:methods_and_metrics}
\head{Compared methods.} The state-of-the-art (SotA) video description methods on ActivityNet Captions include Masked Transformer and Bi-LSTM+TempoAttn~\cite{zhou2018end}. We re-train the models on our dataset splits with the original settings. For a fair comparison, we use exactly the same frame-wise feature from this work for our temporal attention module. For video paragraph description, we compare our methods against the SotA method MFT~\cite{xiong2018move} with the  evaluation script provided by the authors~\cite{xiong2018move}. For image captioning, we compare against two SotA methods, Neural Baby Talk (NBT)~\cite{lu2018neural} and BUTD~\cite{anderson2018bottom}. For a fair comparison, we provide the same region proposal and features for both the baseline BUTD and our method, \ie, from Faster R-CNN pre-trained on Visual Genome (VG). NBT is specially tailored for each dataset (\eg, detector fine-tuning), so we retain the same feature as in the paper, \ie, from ResNet pre-trained on ImageNet. All our experiments are performed three times and the average scores are reported.

\head{Metrics.} To measure the object grounding and attention correctness, we first compute the 
localization accuracy (\textit{Grd.} and \textit{Attn.} in the tables) over GT sentences following~\cite{rohrbach2016grounding, zhou2018weakly}. Given an unseen video, we feed the GT sentence into the model and measure the localization accuracy at each annotated object word. We compare the region with the highest attention weight ($\alpha_i$) or grounding weight ($\beta_j$) against the GT box. An object word is correctly localized if the IoU is over 0.5. 
We also study the attention accuracy on generated sentences, denoted by \textit{$\text{F1}_{all}$} and \textit{$\text{F1}_{loc}$} in the tables. 
In \textit{$\text{F1}_{all}$},
a region prediction is considered correct if the object word is correctly predicted and also correctly localized.
We also compute \textit{$\text{F1}_{loc}$}, which only considers correctly-predicted object words. See Appendix for details.
Due to the sparsity of the annotation, \ie, each object only annotated in one frame, we only consider proposals in the frame of the GT box when computing all the localization accuracies. For the region classification task, we compute the top-1 classification accuracy (\textit{Cls.}~in the tables) for positive regions.
For all metrics, we average the scores across object classes.
To evaluate the sentence quality, we use standard language evaluation metrics, including Bleu@1, Bleu@4, METEOR, CIDEr, and SPICE, and the official evaluation script\footnote{https://github.com/ranjaykrishna/densevid\_eval}.
We additionally perform human evaluation to judge the sentence quality.

\begin{table*}[t]
\centering
{\footnotesize
\begin{tabular}{l|p{0.4cm}p{0.4cm}p{0.4cm}|p{0.6cm}p{0.6cm}p{0.6cm}p{0.6cm}p{0.8cm}|p{0.6cm}p{0.6cm}|p{0.75cm}p{0.75cm}p{0.6cm}}
\toprule
Method & $\lambda_\alpha$ & $\lambda_\beta$ & $\lambda_c$ & B@1 & B@4 & M & C & S & Attn. & Grd. & $\text{F1}_{all}$ & $\text{F1}_{loc}$ & Cls. \\
\midrule
Unsup. (w/o SelfAttn) & 0 & 0 & 0 & 23.2 & 2.28 & 10.9 & 45.6 & \textbf{15.0} & 14.9 & 21.3 & 3.70 & 12.7 & 6.89 \\
Unsup. & 0 & 0 & 0 & 23.0 & 2.27 & 10.7 & 44.6 & 13.8 & 2.42 & 19.7 & 0.28 & 1.13 & 6.06 \\
Sup. Attn. & 0.05 & 0 & 0 & 23.7 & 2.56 & \textbf{11.1} & 47.0 & 14.9 & 34.0 & 37.5 & 6.72 & 22.7 & 0.42 \\
Sup. Grd.  & 0 & 0.5 & 0 & 23.5 & 2.50 & 11.0 & 46.8 & 14.7 & 31.9 & 43.2 & 6.04 & 21.2 & 0.07 \\
Sup. Cls.  & 0 & 0 & 0.1 & 23.3 & 2.43 & 10.9 & 45.7 & 14.1 & 2.59 & 25.8 & 0.35 & 1.43 & \textbf{14.9} \\
Sup. Attn.+Grd. & 0.5 & 0.5 & 0 & \textbf{23.8} & 2.44 & \textbf{11.1} & 46.1 & 14.8 & \textbf{35.1} & 40.6 & 6.79 & 23.0 & 0 \\
Sup. Attn.+Cls. & 0.05 & 0 & 0.1 & \textbf{23.9} & \textbf{2.59} & \textbf{11.2} & \textbf{47.5} & \textbf{15.1} & 34.5 & 41.6 & \textbf{7.11} & \textbf{24.1} & \textbf{14.2} \\
Sup. Grd. +Cls. & 0 & 0.05 & 0.1 & \textbf{23.8} & \textbf{2.59} & \textbf{11.1} & \textbf{47.5} & \textbf{15.0} & 27.1 & \textbf{45.7} & 4.79 & 17.6 & 13.8 \\
Sup. Attn.+Grd.+Cls. & 0.1 & 0.1 & 0.1 & \textbf{23.8} & 2.57 & \textbf{11.1} & 46.9 & \textbf{15.0} & \textbf{35.7} & \textbf{44.9} & \textbf{7.10} & \textbf{23.8} & 12.2 \\
\bottomrule
\end{tabular}
}
\vspace{-5pt}
\caption{Results on \ourdatasetabbrev val set. ``w/o SelfAttn'' indicates self-attention is not used for region feature encoding. Notations: B@1 - Bleu@1, B@4 - Bleu@4, M - METEOR, C - CIDEr, S - SPICE. Attn. and Grd. are the object localization accuracies for attention and grounding on GT sentences. $\text{F1}_{all}$ and $\text{F1}_{loc}$ are the object localization accuracies for attention on generated sentences. Cls. is classification accuracy. All accuracies are in \%. Top two scores on each metric are in bold.}\label{tab:results_vid}
\end{table*}

\begin{table*}[t]
\centering

\begin{subtable}[t]{0.64\textwidth}
\centering
\resizebox{\textwidth}{!}{
\begin{tabular}{l|p{0.6cm}p{0.6cm}p{0.6cm}p{0.6cm}p{0.8cm}|p{0.6cm}p{0.6cm}|p{0.75cm}p{0.75cm}p{0.6cm}}
\toprule
Method & B@1 & B@4 & M & C & S & Attn. & Grd. & $\text{F1}_{all}$ & $\text{F1}_{loc}$ & Cls. \\
\midrule
Masked Transformer~\cite{zhou2018end} & 22.9 & \textbf{2.41} & 10.6 & \textbf{46.1} & 13.7 & -- & -- & -- & -- & -- \\ 
Bi-LSTM+TempoAttn~\cite{zhou2018end} & 22.8 & 2.17 & 10.2 & 42.2 & 11.8 & -- & -- & -- & -- & --\\
\midrule
Our Unsup. (w/o SelfAttn) & 23.1 & 2.16 & 10.8 & 44.9 & \textbf{14.9} & 16.1 & 22.3 & 3.73 & 11.7 & 6.41 \\
Our Sup. Attn.+Cls. (\GVD) & \textbf{23.6} & 2.35 & \textbf{11.0} & 45.5 & 14.7 & \textbf{34.7} & \textbf{43.5} & \textbf{7.59} & \textbf{25.0} & \textbf{14.5} \\
\bottomrule
\end{tabular}
}
\caption{Results on \ourdatasetabbrev test set.}\label{tab:results_vid_test}
\end{subtable}\hspace{30pt}
\begin{subtable}[t]{0.28\textwidth}
\centering
\resizebox{\textwidth}{!}{
    \begin{tabular}{lrr|rr}
    \toprule
    & \multicolumn{2}{c}{vs. Unsupervised} & \multicolumn{2}{c}{vs.~\cite{zhou2018end}} \\ 
    \midrule
    & \multicolumn{2}{c}{Judgments} & \multicolumn{2}{c}{Judgments}\\
        Method &   \% & $\Delta$ &   \% & $\Delta$ \\
        \midrule
        About Equal & 34.9 & & 38.9 &\\
        \cmidrule(rl){2-3}
        \cmidrule(rl){4-5}
        Other is better &29.3 & \multirow{2}{*}{6.5} &27.5 & \multirow{2}{*}{6.1}\\
        \GVD is better &\textbf{35.8} & &\textbf{33.6} &\\
    \bottomrule
    \end{tabular}
    }
    \caption{Human evaluation of sentences.}
    \label{tab:human_eval}
    \vspace{-18pt}
\end{subtable}
\caption{(a) Results on \ourdatasetabbrev test set. The top one score for each metric is in  bold. (b) Human evaluation of sentence quality. We present results for our supervised approach vs. our unsupervised baseline and vs. Masked Transformer~\cite{zhou2018end}.}
\vspace{-10pt}
\end{table*}

\subsection{Implementation Details}\label{sec:imp}
\head{Region proposal and feature.} We uniformly sample 10 frames per video segment (an event in \ourdatasetabbrev) and extract region features. For each frame, we use a Faster R-CNN detector~\cite{ren2015faster} with ResNeXt-101 backbone~\cite{xie2017aggregated} for region proposal and feature extraction (fc6). The detector is pretrained on Visual Genome~\cite{krishna2017visual}. More model and training details are in the Appendix. \\ 
\head{Feature map and attention.} The temporal feature map is essentially a stack of frame-wise appearance and motion features from~\cite{zhou2018end,xiong2016cuhk}. The spatial feature map 
is the conv4 layer output from a ResNet-101~\cite{lu2018neural, he2016deep} model. Note that an average pooling on the temporal or spatial feature map gives the global feature. In video description, we augment the global feature with 
segment positional information (\ie, total number of segments, segment index, start time and end time)
, which is empirically important. 

\head{Hyper-parameters.}
Coefficients $\lambda_\alpha\in\{0.05, 0.1, 0.5\}$, $\lambda_\beta\in\{0.05, 0.1, 0.5\}$, and $\lambda_c\in\{0.1, 0.5, 1\}$ vary in the experiments as a result of model validation. We set $\lambda_\alpha=\lambda_\beta$ when they are both non-zero considering the two losses have a similar functionality. The region encoding size $d=2048$, word embedding size $e=512$ and RNN encoding size $m=1024$ for all methods. Other hyper-parameters in the language module are the same as in~\cite{lu2018neural}. We use a 2-layer 6-head Transformer encoder as the self-attention module~\cite{zhou2018end}. \\

\vspace{-0.1in}
\subsection{Results on \ourdataset}
\subsubsection{Video Event Description}
Although dense video description~\cite{krishna2017visual} further entails localizing the segments to describe on the temporal axis, in this paper we focus on the language generation part and assume the temporal boundaries for events are given. We name this task Video Event Description. 
Results on the validation and test splits of our \ourdataset dataset are shown in Tab.~\ref{tab:results_vid} and Tab.~\ref{tab:results_vid_test}, respectively. Given the selected set of region proposals, the localization upper bound on the val/test sets is 82.5\%/83.4\%, respectively. 

In general, methods with some form of grounding supervision work consistently better than the methods without. Moreover, combining multiple losses, \ie stronger supervision, leads to higher performance.
On the val set, the best variant of supervised methods (\ie, Sup. Attn.+Cls.) ourperforms the best variant of unsupervised methods (\ie, Unsup. (w/o SelfAttn)) by a relative 1-13\% on all the metrics. On the test set, the gaps are small for Bleu@1, METEOR, CIDEr, and SPICE (within $\pm$ 2\%), but the supervised method has a 8.8\% relative improvement on Bleu@4.

The results in  Tab.~\ref{tab:results_vid_test} show that adding box supervision dramatically improves the grounding accuracy from 22.3\% to 43.5\%. Hence, our supervised models can better localize the objects mentioned which can be seen as an improvement in their ability to explain or justify their own description. The attention accuracy also improves greatly on both GT and generated sentences, implying that the supervised models learn to attend on more relevant objects during language generation. However, grounding loss alone fails with respect to classification accuracy (see Tab.~\ref{tab:results_vid}), and therefore the classification loss is required in that case. Conversely, the classification loss alone can implicitly learn grounding and maintains a fair grounding accuracy.

\begin{table*}[t]
\centering
{\footnotesize
\begin{tabular}{l|p{0.3cm}p{0.4cm}|p{0.6cm}p{0.6cm}p{0.6cm}p{0.6cm}p{0.8cm}|p{0.6cm}p{0.6cm}|p{0.75cm}p{0.75cm}p{0.6cm}}
\toprule
Method & VG & Box & B@1 & B@4 & M & C & S & Attn. & Grd. & $\text{F1}_{all}$ & $\text{F1}_{loc}$ & Cls. \\
\midrule
ATT-FCN*~\cite{you2016image} && & 64.7 & 19.9 & 18.5 & -- & -- & -- & -- & -- & -- & -- \\
NBT*~\cite{lu2018neural} & & \checkmark & 69.0 & 27.1 & 21.7 & 57.5 & 15.6 & -- & -- & -- & -- & -- \\
BUTD~\cite{anderson2018bottom} & \checkmark & & 69.4 & \textbf{27.3} & 21.7 & 56.6 & 16.0 & 24.2 & 32.3 & 4.53 & 13.0 & 1.84  \\
\midrule
Our Unsup. (w/o SelfAttn) & \checkmark & & 69.2 & 26.9 & 22.1 & 60.1 & 16.1 & 21.4 & 25.5 & 3.88 & 11.7 & 17.9 \\
Our \GID model & \checkmark  & \checkmark & \textbf{69.9} & \textbf{27.3} & \textbf{22.5} & \textbf{62.3} & \textbf{16.5} & \textbf{41.4} & \textbf{50.9} & \textbf{7.55} & \textbf{22.2} & \textbf{19.2} \\
\bottomrule
\end{tabular}
}
\vspace{-5pt}
\caption{Results on Flickr30k Entities test set. * indicates the results are obtained from the original papers. \GID refers to our Sup. Attn.+Grd.+Cls. model. ``VG'' indicates region features are from VG pre-training. The top one score is in bold.}
\label{tab:results_img_test}
\vspace{-5pt}
\end{table*}

\begin{table}[ht]
    \centering
    {\footnotesize
    \begin{tabular}{l|rrrr}
        \toprule
        Method & B@1 & B@4 & M & C \\
        \midrule
        MFT~\cite{xiong2018move} & 45.5 & 9.78 & 14.6 & 20.4 \\
        Our Unsup. (w/o SelfAttn) & 49.8 & 10.5 & 15.6 & 21.6 \\
        Our GVD & \textbf{49.9} & \textbf{10.7} & \textbf{16.1} & \textbf{22.2} \\ 
        \bottomrule
    \end{tabular}
    }
    \vspace{-5pt}
    \caption{Results of video paragraph description on test set.}\label{tab:results_para}
      \vspace{-5pt}
\end{table}

\head{Comparison to existing methods.} We refer to our best model (Sup. Attn.+Cls.) as \GVD (Grounded Visual Description) and show that it sets the new SotA on ActivityNet Captions for the Bleu@1, METEOR and SPICE metrics, with relative gains of 2.8\%, 3.9\% and 6.8\%, respectively over the previous best~\cite{zhou2018end}. We observe slightly inferior results on Bleu@4 and CIDEr
(-2.8\% and -1.4\%, respectively) but after examining the generated sentences (see Appendix) we see that~\cite{zhou2018end} generates repeated words way more often. This may increase the aforementioned evaluation metrics, but the generated descriptions are of lower quality. 
Another noteworthy observation is that the self-attention context encoder (on top of $\Tilde{R}$) brings consistent improvements on methods with grounding supervision,
but hurts the performance of methods without, \ie, ``Unsup.''. We hypothesize that the extra context and region interaction introduced by the self-attention confuses the region attention module and without any grounding supervision makes it fail to properly attend to the right region, something that leads to a huge attention accuracy drop from 14.9\% to 2.42\%.

\head{Human Evaluation.} Automatic metrics for evaluating generated sentences
have frequently shown to be unreliable and not consistent with human judgments, especially for video description when there is only a single reference \cite{rohrbach17ijcv}.
Hence, we conducted a human evaluation to evaluate the sentence quality on the test set of \ourdataset. 
We randomly sampled 329 video segments and presented the segments and descriptions to the judges. 
From Tab.~\ref{tab:human_eval}, we observe that, while they frequently produce captions with similar quality,  our \GVD works better than the unsupervised baseline (with a significant gap of 6.1\%). 
We can also see that our \GVD approach works better than the Masked Transformer~\cite{zhou2018end} with a significant gap of 6.5\%. We believe these results are a strong indication that our approach is not only better grounded but also generates better sentences, both compared to baselines and prior work \cite{zhou2018end}. 

\subsubsection{Video Paragraph Description}
Besides measuring the quality of each individual description, we also evaluate the coherence among sentences within a video as in~\cite{xiong2018move}. 
We obtained the result file and evaluation script from~\cite{xiong2018move} and evaluated both methods on \textit{our} test split. The results are shown in Tab.~\ref{tab:results_para} and show that we outperform the SotA method of~\cite{xiong2018move} by a large margin. The results are even more surprising given that we generate description for each event separately, without conditioning on previously-generated sentences.
We hypothesize that the temporal attention module can effectively model the event context through the Bi-GRU context encoder and context benefits the coherence of consecutive sentences.

\subsection{Results on Flickr30k Entities}
We show the overall results on image description in Tab.~\ref{tab:results_img_test} (test) and the results on the validation set are in the Appendix. 
The method with the best validation CIDEr score is the full model (Sup. Attn.+Grd.+Cls.), which we further refer to as the \GID model in the table. The upper bounds on the val/test sets are 90.0\%/88.5\%, respectively. 
We see that 
the supervised method outperforms the unsupervised baseline by a relative 1-3.7\% over all the metrics. Our \GID model sets new SotA for all the five metrics with relative gains up to 10\%. In the meantime, object localization and region classification accuracies are significantly boosted, showing that our captions can be better visually explained and understood.

\vspace{-0.1in}
\section{Conclusion}\label{sec:conclude}
In this work, we collected \ourdataset, a novel dataset that allows joint study of video description and grounding. We show how to leverage the noun phrase annotations to generate grounded video descriptions. We also use our dataset to evaluate how well the generated sentences are grounded. We believe our large-scale annotations will also allow for more in-depth analysis which have previously only been able on images, \eg about hallucination \cite{rohrbach18emnlp} and bias \cite{hendricks18eccv}
as well as studying co-reference resolution.
Besides, we showed in our comprehensive experiments on video and image description, how the box supervision can improve the accuracy and the explainability of the generated description by not only generating sentences but also pointing to the corresponding regions in the video frames or image.
According to automatic metrics and human evaluation, on \ourdataset our model performs state-of-the-art \wrt description quality, both when evaluated per sentence or on paragraph level with a significant increase in grounding performance.
We also adapted our model to image description and evaluated it on the Flickr30k Entities dataset where our model outperforms existing methods, both \wrt description quality and grounding accuracy.

\noindent\textbf{Acknowledgement.} The technical work was performed during Luowei's summer internship at Facebook AI Research.
Luowei Zhou and Jason Corso were partly supported by DARPA FA8750-17-2-0125 and NSF IIS 1522904 as part of their affiliation with University of Michigan. This article solely reflects the opinions and conclusions of its authors but not the DARPA or NSF. We thank Chih-Yao Ma for his helpful discussions.

{\small
\bibliographystyle{ieee_fullname}
\bibliography{captioning_grounding,biblioLong}

\begin{thebibliography}{10}\itemsep=-1pt

\bibitem{anderson2018bottom}
Peter Anderson, Xiaodong He, Chris Buehler, Damien Teney, Mark Johnson, Stephen
  Gould, and Lei Zhang.
\newblock Bottom-up and top-down attention for image captioning and visual
  question answering.
\newblock In {\em Proceedings of the IEEE Conference on Computer Vision and
  Pattern Recognition}, pages 6077--6086, 2018.

\bibitem{bigham10uist}
Jeffrey~P. Bigham, Chandrika Jayant, Hanjie Ji, Greg Little, Andrew Miller,
  Robert~C. Miller, Robin Miller, Aubrey Tatarowicz, Brandyn White, Samual
  White, and Tom Yeh.
\newblock Vizwiz: Nearly real-time answers to visual questions.
\newblock In {\em Proceedings of the 23Nd Annual ACM Symposium on User
  Interface Software and Technology}, UIST '10, pages 333--342, New York, NY,
  USA, 2010. ACM.

\bibitem{caba2015activitynet}
Fabian Caba~Heilbron, Victor Escorcia, Bernard Ghanem, and Juan Carlos~Niebles.
\newblock Activitynet: A large-scale video benchmark for human activity
  understanding.
\newblock In {\em Proceedings of the IEEE Conference on Computer Vision and
  Pattern Recognition}, pages 961--970, 2015.

\bibitem{das2017human}
Abhishek Das, Harsh Agrawal, Larry Zitnick, Devi Parikh, and Dhruv Batra.
\newblock Human attention in visual question answering: Do humans and deep
  networks look at the same regions?
\newblock {\em Computer Vision and Image Understanding}, 163:90--100, 2017.

\bibitem{das2013thousand}
Pradipto Das, Chenliang Xu, Richard~F Doell, and Jason~J Corso.
\newblock A thousand frames in just a few words: Lingual description of videos
  through latent topics and sparse object stitching.
\newblock In {\em Proceedings of the IEEE conference on computer vision and
  pattern recognition}, pages 2634--2641, 2013.

\bibitem{gu2017ava}
Chunhui Gu, Chen Sun, Sudheendra Vijayanarasimhan, Caroline Pantofaru, David~A
  Ross, George Toderici, Yeqing Li, Susanna Ricco, Rahul Sukthankar, Cordelia
  Schmid, et~al.
\newblock Ava: A video dataset of spatio-temporally localized atomic visual
  actions.
\newblock In {\em Proceedings of the IEEE international conference on computer
  vision}, 2018.

\bibitem{he2017mask}
Kaiming He, Georgia Gkioxari, Piotr Doll{\'a}r, and Ross Girshick.
\newblock Mask r-cnn.
\newblock In {\em Computer Vision (ICCV), 2017 IEEE International Conference
  on}, pages 2980--2988. IEEE, 2017.

\bibitem{he2016deep}
Kaiming He, Xiangyu Zhang, Shaoqing Ren, and Jian Sun.
\newblock Deep residual learning for image recognition.
\newblock In {\em Proceedings of the IEEE conference on computer vision and
  pattern recognition}, pages 770--778, 2016.

\bibitem{hendricks18eccv}
Lisa~Anne Hendricks, Kaylee Burns, Kate Saenko, Trevor Darrell, and Anna
  Rohrbach.
\newblock Women also snowboard: Overcoming bias in captioning models.
\newblock In {\em Proceedings of the European Conference on Computer Vision},
  pages 771--787, 2018.

\bibitem{jiang2018pythia}
Yu Jiang, Vivek Natarajan, Xinlei Chen, Marcus Rohrbach, Dhruv Batra, and Devi
  Parikh.
\newblock Pythia v0. 1: the winning entry to the vqa challenge 2018.
\newblock {\em arXiv preprint arXiv:1807.09956}, 2018.

\bibitem{krishna2017dense}
Ranjay Krishna, Kenji Hata, Frederic Ren, Li Fei-Fei, and Juan~Carlos Niebles.
\newblock Dense-captioning events in videos.
\newblock In {\em Proceedings of the IEEE International Conference on Computer
  Vision (ICCV)}, pages 706--715, 2017.

\bibitem{krishna2017visual}
Ranjay Krishna, Yuke Zhu, Oliver Groth, Justin Johnson, Kenji Hata, Joshua
  Kravitz, Stephanie Chen, Yannis Kalantidis, Li-Jia Li, David~A Shamma, et~al.
\newblock Visual genome: Connecting language and vision using crowdsourced
  dense image annotations.
\newblock {\em International Journal of Computer Vision}, 123(1):32--73, 2017.

\bibitem{kulkarni2013babytalk}
Girish Kulkarni, Visruth Premraj, Vicente Ordonez, Sagnik Dhar, Siming Li,
  Yejin Choi, Alexander~C Berg, and Tamara~L Berg.
\newblock Babytalk: Understanding and generating simple image descriptions.
\newblock {\em IEEE Transactions on Pattern Analysis and Machine Intelligence},
  35(12):2891--2903, 2013.

\bibitem{li2018jointly}
Yehao Li, Ting Yao, Yingwei Pan, Hongyang Chao, and Tao Mei.
\newblock Jointly localizing and describing events for dense video captioning.
\newblock In {\em Proceedings of the IEEE Conference on Computer Vision and
  Pattern Recognition}, pages 7492--7500, 2018.

\bibitem{liu2017attention}
Chenxi Liu, Junhua Mao, Fei Sha, and Alan~L Yuille.
\newblock Attention correctness in neural image captioning.
\newblock In {\em Proceedings of the Conference on Artificial Intelligence
  (AAAI)}, pages 4176--4182, 2017.

\bibitem{lu2018neural}
Jiasen Lu, Jianwei Yang, Dhruv Batra, and Devi Parikh.
\newblock Neural baby talk.
\newblock In {\em Proceedings of the IEEE Conference on Computer Vision and
  Pattern Recognition}, pages 7219--7228, 2018.

\bibitem{ma2018attend}
Chih-Yao Ma, Asim Kadav, Iain Melvin, Zsolt Kira, Ghassan AlRegib, and Hans
  Peter~Graf.
\newblock Attend and interact: Higher-order object interactions for video
  understanding.
\newblock In {\em Proceedings of the IEEE Conference on Computer Vision and
  Pattern Recognition}, pages 6790--6800, 2018.

\bibitem{manning2014stanford}
Christopher Manning, Mihai Surdeanu, John Bauer, Jenny Finkel, Steven Bethard,
  and David McClosky.
\newblock The stanford corenlp natural language processing toolkit.
\newblock In {\em Proceedings of 52nd annual meeting of the association for
  computational linguistics: system demonstrations}, pages 55--60, 2014.

\bibitem{mitchell2012midge}
Margaret Mitchell, Xufeng Han, Jesse Dodge, Alyssa Mensch, Amit Goyal, Alex
  Berg, Kota Yamaguchi, Tamara Berg, Karl Stratos, and Hal Daum{\'e}~III.
\newblock Midge: Generating image descriptions from computer vision detections.
\newblock In {\em Proceedings of the 13th Conference of the European Chapter of
  the Association for Computational Linguistics}, pages 747--756. Association
  for Computational Linguistics, 2012.

\bibitem{pan2017video}
Yingwei Pan, Ting Yao, Houqiang Li, and Tao Mei.
\newblock Video captioning with transferred semantic attributes.
\newblock In {\em Proceedings of the IEEE Conference on Computer Vision and
  Pattern Recognition (CVPR)}, volume~2, page~3, 2017.

\bibitem{park18cvpr}
Dong~Huk Park, Lisa~Anne Hendricks, Zeynep Akata, Anna Rohrbach, Bernt Schiele,
  Trevor Darrell, and Marcus Rohrbach.
\newblock Multimodal explanations: Justifying decisions and pointing to the
  evidence.
\newblock In {\em Proceedings of the IEEE Conference on Computer Vision and
  Pattern Recognition (CVPR)}, 2018.

\bibitem{pennington2014glove}
Jeffrey Pennington, Richard Socher, and Christopher Manning.
\newblock Glove: Global vectors for word representation.
\newblock In {\em Proceedings of the 2014 conference on empirical methods in
  natural language processing (EMNLP)}, pages 1532--1543, 2014.

\bibitem{plummer2015flickr30k}
Bryan~A Plummer, Liwei Wang, Chris~M Cervantes, Juan~C Caicedo, Julia
  Hockenmaier, and Svetlana Lazebnik.
\newblock Flickr30k entities: Collecting region-to-phrase correspondences for
  richer image-to-sentence models.
\newblock In {\em Proceedings of the IEEE international conference on computer
  vision}, pages 2641--2649, 2015.

\bibitem{ren2015faster}
Shaoqing Ren, Kaiming He, Ross Girshick, and Jian Sun.
\newblock Faster r-cnn: Towards real-time object detection with region proposal
  networks.
\newblock In {\em Advances in neural information processing systems}, pages
  91--99, 2015.

\bibitem{rohrbach18emnlp}
Anna Rohrbach, Lisa~Anne Hendricks, Kaylee Burns, Trevor Darrell, and Kate
  Saenko.
\newblock Object hallucination in image captioning.
\newblock In {\em Proceedings of the 2018 Conference on Empirical Methods in
  Natural Language Processing}, pages 4035--4045, 2018.

\bibitem{rohrbach2016grounding}
Anna Rohrbach, Marcus Rohrbach, Ronghang Hu, Trevor Darrell, and Bernt Schiele.
\newblock Grounding of textual phrases in images by reconstruction.
\newblock In {\em European Conference on Computer Vision}, pages 817--834.
  Springer, 2016.

\bibitem{rohrbach2017generating}
Anna Rohrbach, Marcus Rohrbach, Siyu Tang, Seong~Joon Oh, and Bernt Schiele.
\newblock Generating descriptions with grounded and co-referenced people.
\newblock In {\em Proceedings of the IEEE Conference on Computer Vision and
  Pattern Recognition (CVPR)}, 2017.

\bibitem{rohrbach17ijcv}
Anna Rohrbach, Atousa Torabi, Marcus Rohrbach, Niket Tandon, Chris Pal, Hugo
  Larochelle, Aaron Courville, and Bernt Schiele.
\newblock Movie description.
\newblock {\em International Journal of Computer Vision (IJCV)}, 2017.

\bibitem{vaswani2017attention}
Ashish Vaswani, Noam Shazeer, Niki Parmar, Jakob Uszkoreit, Llion Jones,
  Aidan~N Gomez, {\L}ukasz Kaiser, and Illia Polosukhin.
\newblock Attention is all you need.
\newblock In {\em Advances in Neural Information Processing Systems}, pages
  5998--6008, 2017.

\bibitem{xie2017aggregated}
Saining Xie, Ross Girshick, Piotr Doll{\'a}r, Zhuowen Tu, and Kaiming He.
\newblock Aggregated residual transformations for deep neural networks.
\newblock In {\em Computer Vision and Pattern Recognition (CVPR), 2017 IEEE
  Conference on}, pages 5987--5995. IEEE, 2017.

\bibitem{xiong2018move}
Yilei Xiong, Bo Dai, and Dahua Lin.
\newblock Move forward and tell: A progressive generator of video descriptions.
\newblock {\em Proceedings of the European Conference on Computer Vision},
  2018.

\bibitem{xiong2016cuhk}
Yuanjun Xiong, Limin Wang, Zhe Wang, Bowen Zhang, Hang Song, Wei Li, Dahua Lin,
  Yu Qiao, Luc Van~Gool, and Xiaoou Tang.
\newblock Cuhk \& ethz \& siat submission to activitynet challenge 2016.
\newblock {\em arXiv preprint arXiv:1608.00797}, 2016.

\bibitem{xu2015show}
Kelvin Xu, Jimmy Ba, Ryan Kiros, Kyunghyun Cho, Aaron Courville, Ruslan
  Salakhudinov, Rich Zemel, and Yoshua Bengio.
\newblock Show, attend and tell: Neural image caption generation with visual
  attention.
\newblock In {\em International conference on machine learning}, pages
  2048--2057, 2015.

\bibitem{yamaguchi2017spatio}
Masataka Yamaguchi, Kuniaki Saito, Yoshitaka Ushiku, and Tatsuya Harada.
\newblock Spatio-temporal person retrieval via natural language queries.
\newblock In {\em Proceedings of the IEEE International Conference on Computer
  Vision}, pages 1453--1462, 2017.

\bibitem{yao2015describing}
Li Yao, Atousa Torabi, Kyunghyun Cho, Nicolas Ballas, Christopher Pal, Hugo
  Larochelle, and Aaron Courville.
\newblock Describing videos by exploiting temporal structure.
\newblock In {\em Proceedings of the IEEE international conference on computer
  vision}, pages 4507--4515, 2015.

\bibitem{yao2017boosting}
Ting Yao, Yingwei Pan, Yehao Li, Zhaofan Qiu, and Tao Mei.
\newblock Boosting image captioning with attributes.
\newblock In {\em IEEE International Conference on Computer Vision, ICCV},
  pages 22--29, 2017.

\bibitem{you2016image}
Quanzeng You, Hailin Jin, Zhaowen Wang, Chen Fang, and Jiebo Luo.
\newblock Image captioning with semantic attention.
\newblock In {\em Proceedings of the IEEE conference on computer vision and
  pattern recognition}, pages 4651--4659, 2016.

\bibitem{yu2017supervising}
Youngjae Yu, Jongwook Choi, Yeonhwa Kim, Kyung Yoo, Sang-Hun Lee, and Gunhee
  Kim.
\newblock Supervising neural attention models for video captioning by human
  gaze data.
\newblock In {\em IEEE Conference on Computer Vision and Pattern Recognition
  (CVPR 2017). Honolulu, Hawaii}, pages 2680--29, 2017.

\bibitem{zanfir16accv}
Mihai Zanfir, Elisabeta Marinoiu, and Cristian Sminchisescu.
\newblock Spatio-temporal attention models for grounded video captioning.
\newblock In {\em Asian Conference on Computer Vision}, pages 104--119, 2016.

\bibitem{zhang2019interpretable}
Yundong Zhang, Juan~Carlos Niebles, and Alvaro Soto.
\newblock Interpretable visual question answering by visual grounding from
  attention supervision mining.
\newblock In {\em 2019 IEEE Winter Conference on Applications of Computer
  Vision (WACV)}, pages 349--357. IEEE, 2019.

\bibitem{zhou2018weakly}
Luowei Zhou, Nathan Louis, and Jason~J Corso.
\newblock Weakly-supervised video object grounding from text by loss weighting
  and object interaction.
\newblock {\em Proceedings of the British Machine Vision Conference (BMVC)},
  2018.

\bibitem{zhou2017watch}
Luowei Zhou, Chenliang Xu, Parker Koch, and Jason~J Corso.
\newblock Watch what you just said: Image captioning with text-conditional
  attention.
\newblock In {\em Proceedings of the on Thematic Workshops of ACM Multimedia
  2017}, pages 305--313. ACM, 2017.

\bibitem{zhou2018end}
Luowei Zhou, Yingbo Zhou, Jason~J Corso, Richard Socher, and Caiming Xiong.
\newblock End-to-end dense video captioning with masked transformer.
\newblock In {\em Proceedings of the IEEE Conference on Computer Vision and
  Pattern Recognition}, pages 8739--8748, 2018.

\end{thebibliography}
}

\newpage

\clearpage

\appendix
\section{Appendix}\label{sec:app}
This Appendix provides additional details, evaluations, and qualitative results.
\begin{itemize}
    \item In Sec.~\ref{sec:dataset}, we provide more details on our dataset including the annotation interface and examples of our dataset, which are shown in Figs.~\ref{fig:examples},~\ref{fig:screenshot}.
    \item In Sec.~\ref{sec:metric_details}, we clarify on the four localization metrics.
    \item In Sec.~\ref{sec:results:activitynet}, we provide additional ablations and results on our \ourdataset dataset, including qualitative results, which are shown in Figs.~\ref{fig:anet_vis1},~\ref{fig:anet_vis2}.
    \item In Sec.~\ref{sec:results:flickr30k}, we provide additional results on the Flickr30kEntities dataset, including qualitative results, which are shown in Fig.~\ref{fig:flickr_vis}.
    \item In Sec.~\ref{sec:implementation:details}, we provide more implementation details (\eg, training details).
\end{itemize}

\subsection{Dataset}
\label{sec:dataset}

\head{Definition of a noun phrase}. Following the convention from Flickr30k Entities dataset~\cite{plummer2015flickr30k}, we define noun phrase as:
\begin{itemize}
\item short (avg. 2.23 words), non-recursive phrases (\eg, the complex NP ``the man in a white shirt with a heart'' is split into three: ``the man'', ``a white shirt'', and ``a heart'')
\item refer to a specific region in the image so as to be annotated as a bounding box.
\item could be
	\begin{itemize}
    \item a single instance (\eg, a cat),
    \item multiple distinct instances (\eg two men),
    \item a group of instances (\eg, a group of people),
    \item a region or scene (\eg, grass/field/kitchen/town),
    \item a pronoun, \eg, it, him, they.
    \end{itemize}
\item could include
	\begin{itemize}
	\item adjectives (\eg, a \textit{white} shirt),
    \item determiners (\eg, \textit{A} piece of exercise equipment),
    \item prepositions (\eg the woman \textit{on the right})
    \item other noun phrases, if they refer to the identical bounding concept \& bounding box (\eg, a group of people, a shirt of red color)
    \end{itemize}
\end{itemize}

\head{Annotator instructions}

\begin{figure*}[p]
\centering
\begin{minipage}[t]{0.49\textwidth}
\centering
\includegraphics[width=\textwidth]{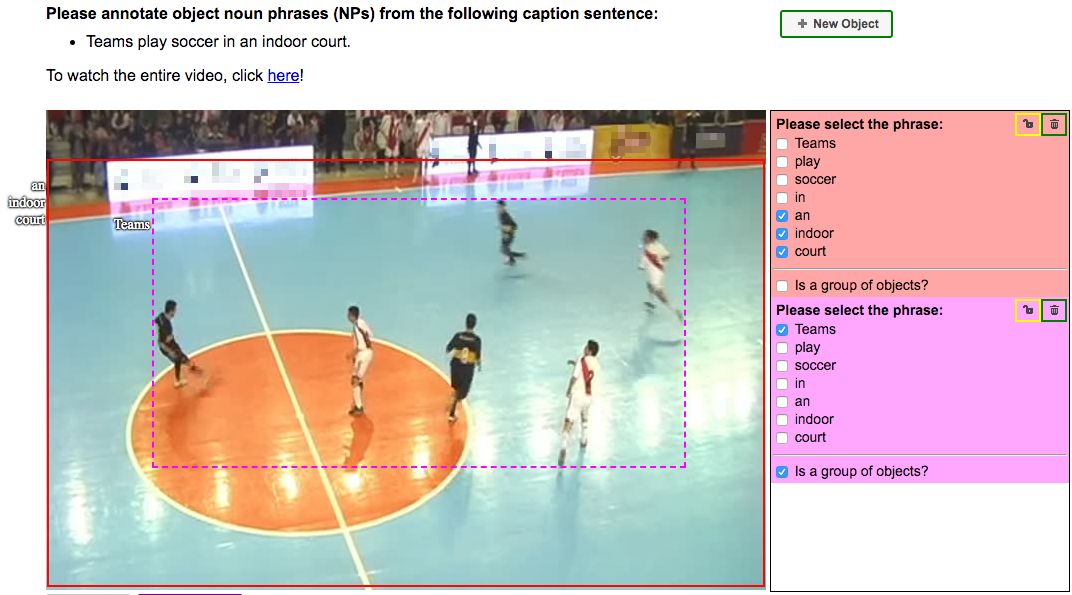} \\ (a) ``Teams'' refers to more than 5 instances and hence should be annotated as a group.
\end{minipage}
\begin{minipage}[t]{0.49\textwidth}
\centering
\includegraphics[width=\textwidth]{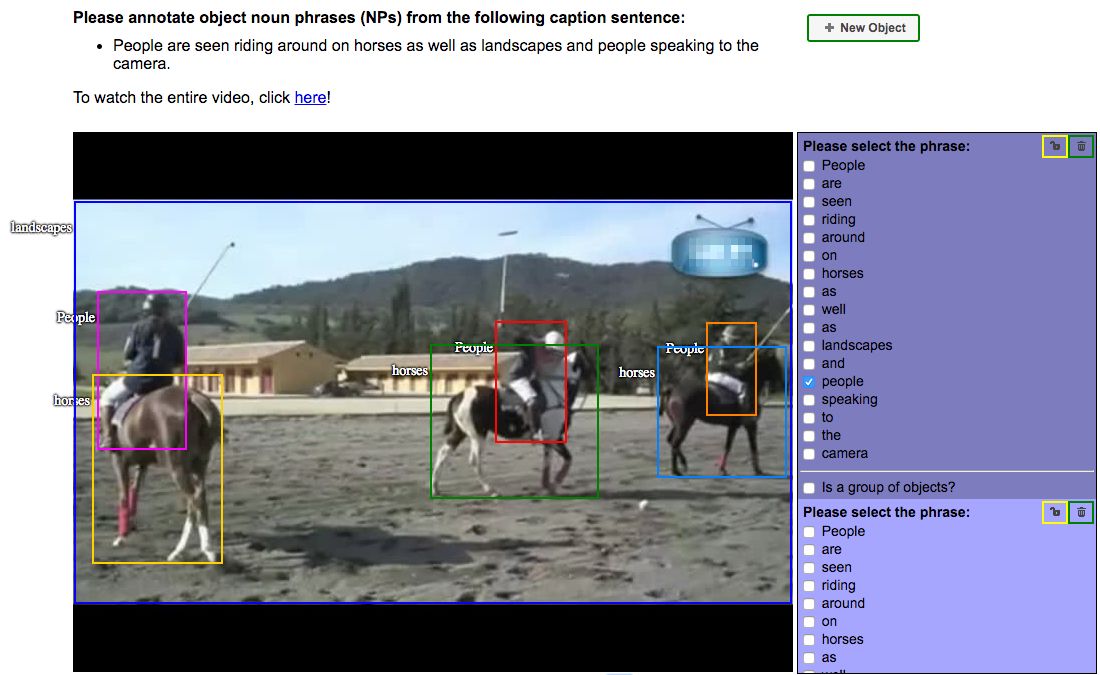} \\ (b) ``People'' and ``horses'' can be clearly separated and the \# of instances each is $\leq$ 5. So, annotate them all.
\end{minipage}
\begin{minipage}[t]{0.49\textwidth}
\centering
\includegraphics[width=\textwidth]{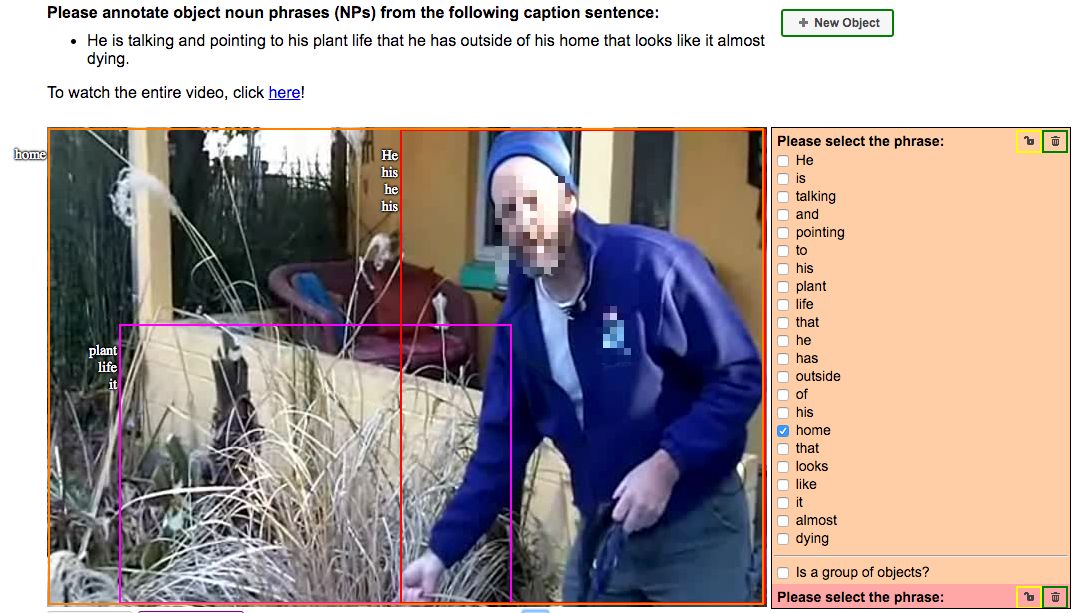} \\  (c) ``plant life'' and ``it'' refer to the same box and ``He'', ``'his'', ``he'', ``his'' all refer to the same box.
\end{minipage}
\begin{minipage}[t]{0.49\textwidth}
\centering
\includegraphics[width=\textwidth]{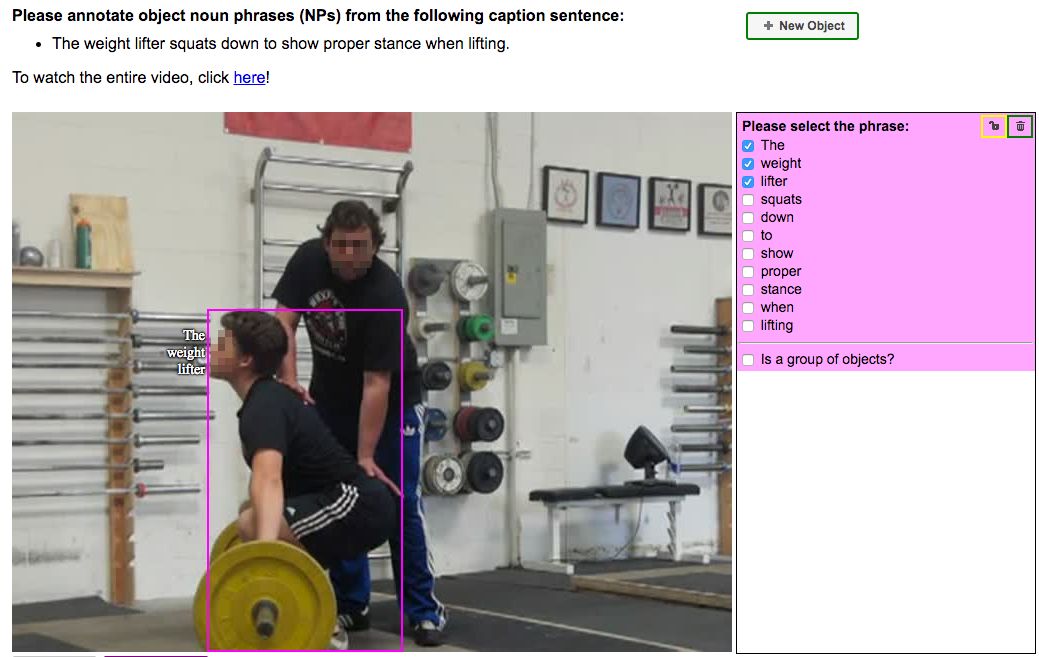} \\ (d) Only annotate the NP mentioned in the sentence, in this case, ``The weight lifter''. ``proper stance'' is a NP but not annotated because it is abstract/not an object in the scene.
\end{minipage}
\begin{minipage}[t]{0.48\textwidth}
\centering
\includegraphics[width=\textwidth]{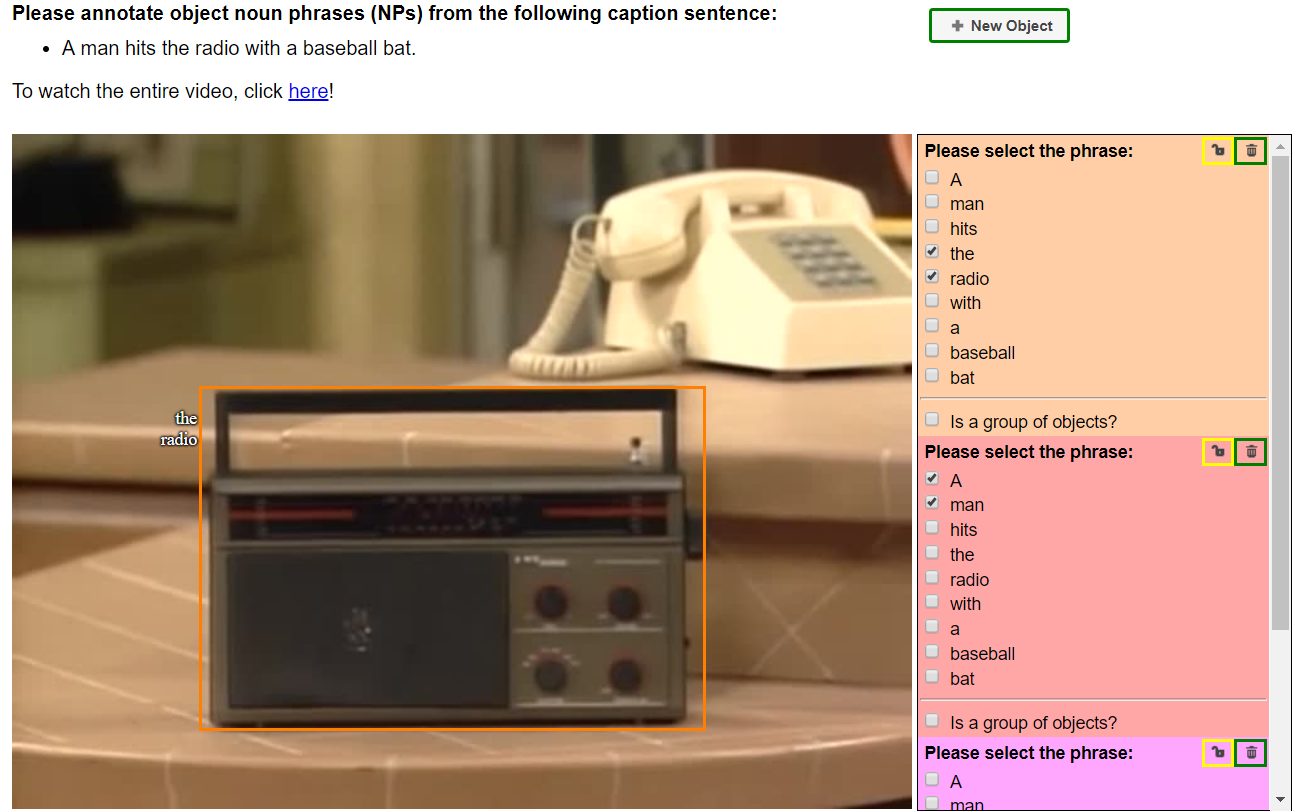} \\  (e) Note that (e) and (f) refer to the same video segment. See the caption of (f) for more details.
\end{minipage}
\begin{minipage}[t]{0.48\textwidth}
\centering
\includegraphics[width=\textwidth]{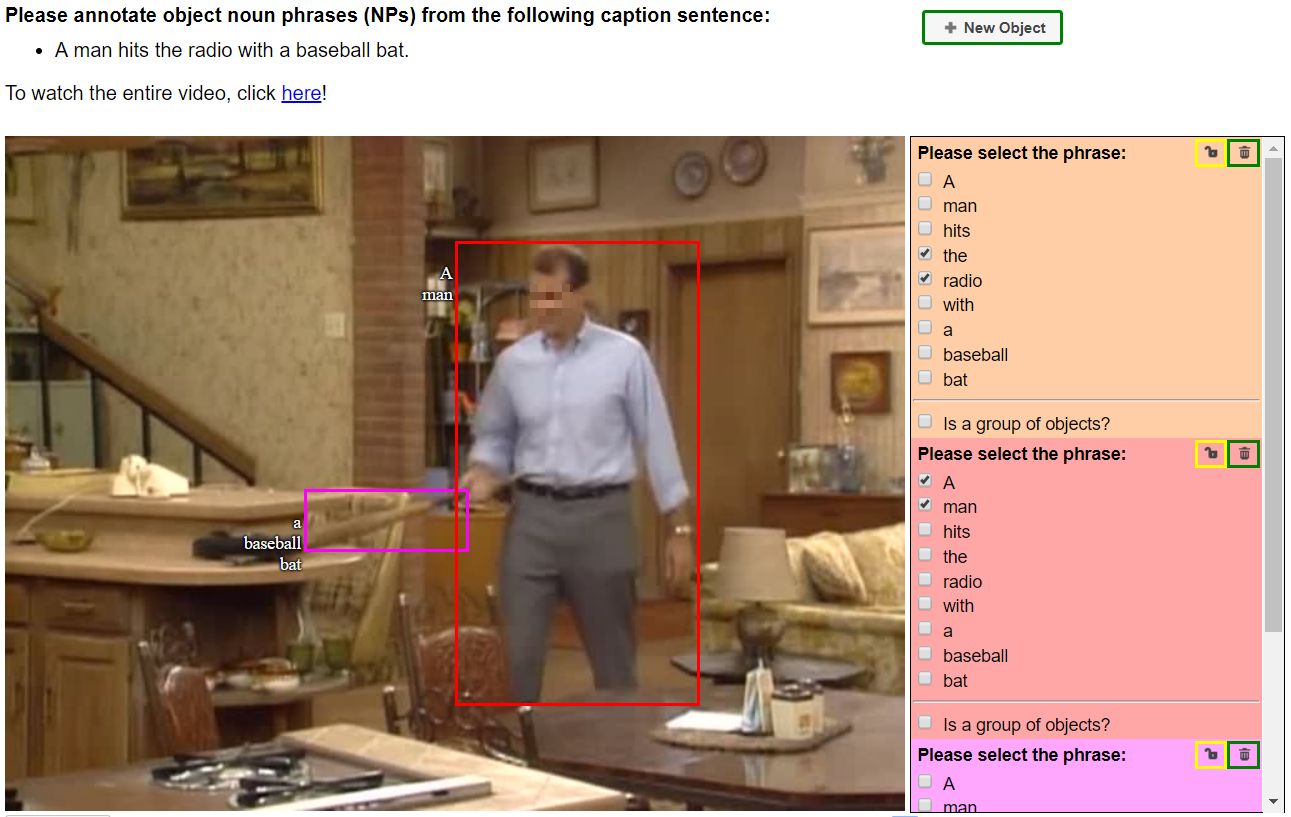} \\ (f) ``The radio'' is annotated in a different frame as ``a man'' and ``a baseball bat'', since it cannot be clearly observed in the same frame. 
\end{minipage}
\caption{Examples of our \ourdataset annotations in the annotation interface.}
\label{fig:examples}
\end{figure*}

Further instructions include:
\begin{itemize}
\item Each word from the caption can appear in at most one NP. ``A man in a white shirt'' and ``a white shirt'' should not be annotated at the same time.
\item Annotate multiple boxes for the same NP if the NP refers to multiple instances.
	\begin{itemize}
	\item If there are more than 5 instances/boxes (\eg, six cats or many young children), mark all instances as a single box and mark as ``a group of objects''. 
    \item Annotate 5 or fewer instances with a single box if the instances are difficult to separate, \eg if they are strongly occluding each other.
    \end{itemize}
\item We don't annotate a NP if it's abstract or not presented in the scene (\eg, ``the camera'' in ``A man is speaking to the camera'')
\item One box can correspond to multiple NPs in the sentence (\eg, ``the man'' and ``him''), \ie, we annotate co-references within one sentence.
\end{itemize}
See Fig.~\ref{fig:examples} for more examples.

\begin{figure*}[t]
\centering
   \includegraphics[width=1\linewidth]{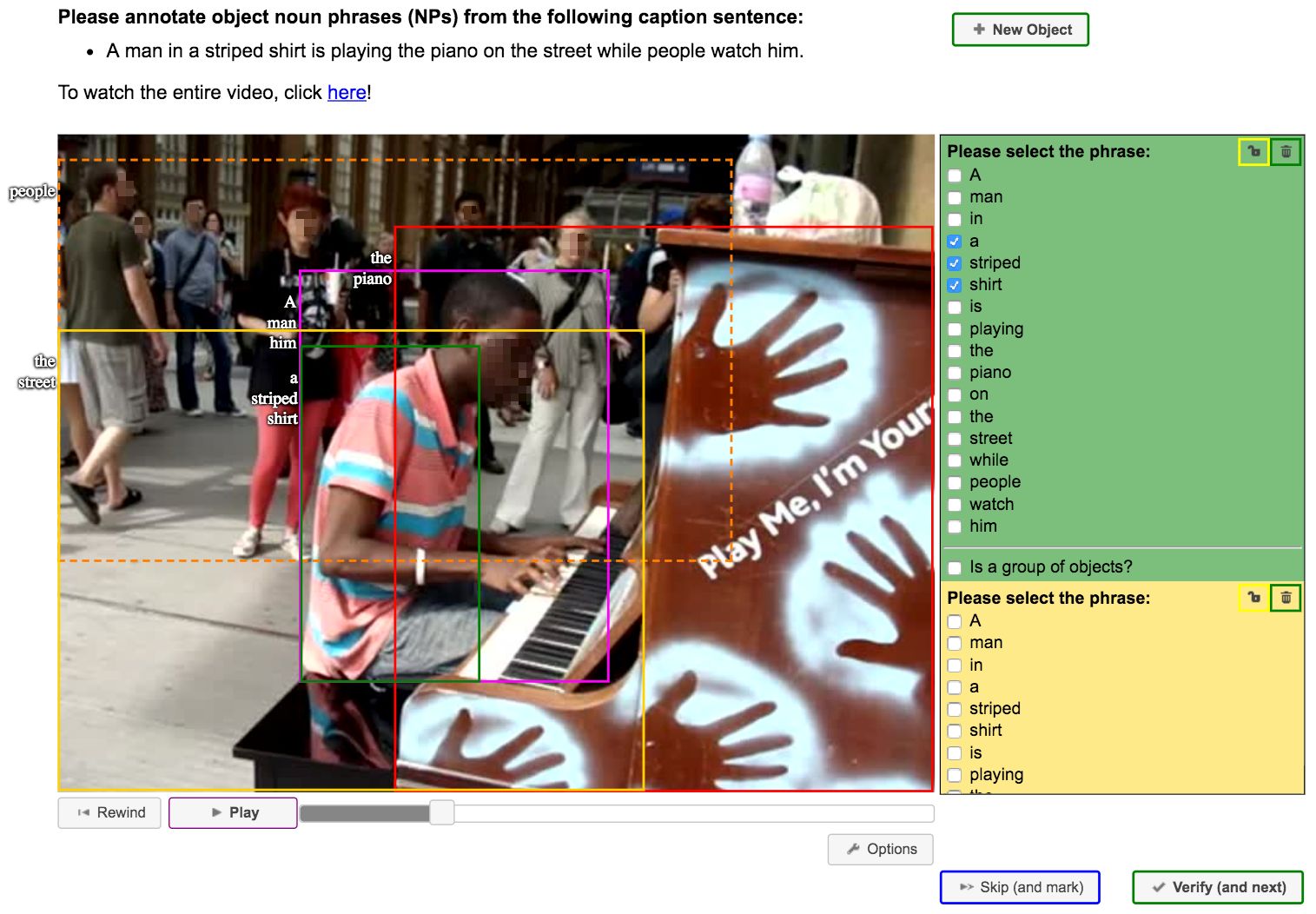}
   \caption{
   A screen shot of our annotation interface. The ``verify (and next)'' button indicates the annotation is under the verification mode, where the initial annotation is loaded and could be revised.
   }
\label{fig:screenshot}
\end{figure*}

\head{Annotation interface.} We show a screen shot of the interface in Fig.~\ref{fig:screenshot}.

\head{Validation process.} We deployed a rigid quality control process during annotations.
We were in daily contact with the annotators, encouraged them to flag all examples that were unclear and inspected a sample of the annotations daily, providing them with feedback on possible spotted annotation errors or guideline violations. We also had a post-annotation verification process where all the annotations are verified by human annotators.

\head{Dataset statistics.} The average number of annotated boxes per video segment is 2.56 and the standard deviation is 2.04. The average number of object labels per box is 1.17 and the standard deviation is 0.47. The top ten frequent objects are ``man'', ``he'', ``people'', ``they'', ``she'', ``woman'', ``girl'', ``person'', ``it'', and ``boy''. Note that the statistics are on object boxes, \ie, after pre-processing. \\

\head{List of objects}. Tab. 10 lists all the 432 object classes which we use in our approach. We threshold at  50 occurrences. Note that the annotations in \ourdataset also contain the full noun phrases w/o thresholds. \\

\subsection{Localization Metrics}
\label{sec:metric_details}
We use four localization metrics, $Attn.$, $Grd.$, $F1_{all}$, and $F1_{loc}$ as mentioned in Sec.~\ref{sec:methods_and_metrics}. The first two are computed on the GT sentences, \ie, during inference, we feed the GT sentences into the model and compute the attention and grounding localization accuracies. The last two measure are computed on the generated sentences, \ie, given a test video segment, we perform the standard language generation inference and compute attention localization accuracy (no grounding measurement here because it is usually evaluated on GT sentences). We define  \textit{$\text{F1}_{all}$} and \textit{$\text{F1}_{loc}$} as follows.

We define the number of object words in the generated sentences as $A$, the number of object words in the GT sentences as $B$, the number of correctly predicted object words in the generated sentences as $C$ and the counterpart in the GT sentences as $D$, and the number of correctly predicted and localized words as $E$. A region prediction is considered correct if the object word is correctly predicted and also correctly localized (\ie, IoU with GT box $>$ 0.5).

In \textit{$\text{F1}_{all}$}, the precision and recall can be defined as:

\begin{align}
    \text{Precision}_{all}  = \frac{E}{A}, \;\;\;\;
    \text{Recall}_{all}  = \frac{E}{B}
\end{align}

\noindent However, since having box annotation for every single object in the scene is unlikely, an incorrectly-predicted word might not necessarily be a hallucinated object. 
Hence, we also compute \textit{$\text{F1}_{loc}$}, which only considers correctly-predicted object words, i.e., only measures the localization quality and ignores errors result from the language generation. The precision and recall for \textit{$\text{F1}_{loc}$} are defined as:

\begin{align}
    \text{Precision}_{loc}  = \frac{E}{C}, \;\;\;\;
    \text{Recall}_{loc}  = \frac{E}{D}
\end{align}

If multiple instances of the same object exist in the target sentence, we only consider the first instance.
The precision and recall for the two metrics are computed for each object class, but it is set to zero if an object class has never been predicted. Finally, we average the scores by dividing by the total number of object classes in a particular split (val or test). 

During model training, we restrict the grounding region candidates within the target frame (w/ GT box), \ie, only consider the $N_f$ proposals on the frame $f$ with the GT box.

\subsection{Results on \ourdataset}
\label{sec:results:activitynet}
We first include here the precision and recall associated with \textit{$\text{F1}_{all}$} and \textit{$\text{F1}_{loc}$} (see Tabs.~\ref{tab:anet_pr_val}, \ref{tab:anet_pr_test}).

\begin{table}[t]
\centering
{\footnotesize
    \begin{tabular}{l|rr|rr}
    \toprule
    & \multicolumn{2}{c}{$\text{F1}_{all}$} & \multicolumn{2}{c}{$\text{F1}_{loc}$}\\
        Method &   Precision & Recall & Precision & Recall \\
        \midrule
        Unsup. (w/o SelfAttn) & 3.76 & 3.63 & 12.6 & 12.9 \\
        Unsup.  & 0.28 & 0.27 & 1.13 & 1.13 \\
        Sup. Attn.  & 6.71 & 6.73 & 22.6 & 22.8 \\
        Sup. Grd.   & 6.25 & 5.84 & 21.2 & 21.2 \\
        Sup. Cls.   & 0.40 & 0.32 & 1.39 & 1.47 \\
        Sup. Attn.+Grd.  & 7.07 & 6.54 & 23.0 & 23.0 \\
        Sup. Attn.+Cls.  & 7.29 & 6.94 & 24.0 & 24.1 \\
        Sup. Grd. +Cls.  & 4.94 & 4.64 & 17.7  & 17.6 \\
        Sup. Attn.+Grd.+Cls.  & 7.42 & 6.81 & 23.7 & 23.9 \\
        \bottomrule
    \end{tabular}\vspace{-5pt}
    \caption{Attention precision and recall on generated sentences on \ourdatasetabbrev val set. All values are in \%.}\vspace{-5pt}
    \label{tab:anet_pr_val}
}
\end{table}

\begin{table}[t]
\centering
{\footnotesize
    \begin{tabular}{l|rr|rr}
    \toprule
    & \multicolumn{2}{c}{$\text{F1}_{all}$} & \multicolumn{2}{c}{$\text{F1}_{loc}$}\\
        Method &   Precision & Recall & Precision & Recall \\
        \midrule
        Unsup. (w/o SelfAttn) & 3.62 & 3.85 & 11.7 & 11.8 \\
        Sup. Attn.+Cls. & 7.64 & 7.55 & 25.1 & 24.8 \\
        \bottomrule
    \end{tabular}\vspace{-5pt}
    \caption{Attention precision and recall on generated sentences on \ourdatasetabbrev test set. All values are in \%.}\vspace{-5pt}
    \label{tab:anet_pr_test}
}
\end{table}

\head{Temporal attention \& region attention.}
We conduct ablation studies on the two attention modules to study the impact of each component on the overall performance (see Tab.~\ref{tab:results_attn}). Each module alone performs similarly and the combination of two performs the best, which indicates the two attention modules are complementary. We hypothesize that the temporal attention captures the coarse-level details while the region attention captures more fine-grained details.
Note that the region attention module takes in a lower sampling rate input than the temporal attention module, so we expect it can be further improved if having a higher sampling rate and the context (other events in the video). We leave this for future studies.

\begin{table}[t]
    \centering
    {\footnotesize
    \begin{tabular}{l|rrrrr}
        \toprule
        Method & B@1 & B@4 & M & C & S \\
        \midrule
        Region Attn. & 23.2 & 2.55 & 10.9 & 43.5 & 14.5 \\ 
        Tempo. Attn. & 23.5 & 2.45 & 11.0 & 44.3 & 14.0 \\
        Both & \textbf{23.9} & \textbf{2.59} & \textbf{11.2} & \textbf{47.5} & \textbf{15.1} \\ 
        \bottomrule
    \end{tabular}
    }\vspace{-5pt}
      \caption{Ablation study for two attention modules using our best model. Results reported on val set.}\vspace{-10pt}
      \label{tab:results_attn}
\end{table}

\head{Notes on Video Paragraph Description.} The authors of the SoTA method~\cite{xiong2018move} kindly provided us with their result file and evaluation script, but as they were unable to provide us with
their splits, we evaluated both methods on \textit{our} test split. Even though we are under an unfair disadvantage, \ie, the authors' val split might contain videos from our test split, we still outperform SotA method by a large margin, with relative improvements of 8.9-10\% on all the metrics (as shown in Tab.~\ref{tab:results_para}).

\begin{table*}[t]
\centering
{\footnotesize
\begin{tabular}{l|p{0.4cm}p{0.4cm}p{0.4cm}|p{0.6cm}p{0.6cm}p{0.6cm}p{0.6cm}p{0.8cm}|p{0.6cm}p{0.6cm}|p{0.75cm}p{0.75cm}p{0.6cm}}
\toprule
Method & $\lambda_\alpha$ & $\lambda_\beta$ & $\lambda_c$ & B@1 & B@4 & M & C & S & Attn. & Grd. & $\text{F1}_{all}$ & $\text{F1}_{loc}$ & Cls. \\
\midrule
Unsup. (w/o SelfAttn) & 0 & 0 & 0 & 70.0 & 27.5 & 22.0 & 60.4 & 15.9 & 22.0 & 25.9 & 4.44 & 12.8 & 17.6  \\
Unsup. & 0 & 0 & 0 & 69.3 & 26.8 & 22.1 & 59.4 & 15.7 & 4.04 & 16.3 & 0.80 & 2.09 & 1.35 \\
Sup. Attn. & 0.1 & 0 & 0 & \textbf{71.0} & \textbf{28.2} & \textbf{22.7} & 63.0 & \textbf{16.3} & \textbf{42.3} & 44.1 & 8.08 & 22.4 & 6.59  \\
Sup. Grd.  & 0 & 0.1 & 0 & 70.1 & 27.6 & 22.5 & \textbf{63.1} & 16.1 & 38.5 & 49.5 & 7.59 & 21.0 & 0.03 \\
Sup. Cls. (w/o SelfAttn)  & 0 & 0 & 1 & 70.1  & 27.6 & 22.0 & 60.2 & 15.8 & 20.9 & 32.1 & 4.12 & 11.5 & \textbf{19.9} \\
Sup. Attn.+Grd. & 0.1 & 0.1 & 0 & 70.2 & 27.6 & 22.5 & 62.3 & \textbf{16.3} & \textbf{42.7} & 49.8 & \textbf{8.62} & \textbf{23.6} & 0 \\
Sup. Attn.+Cls. & 0.1 & 0 & 1 & 70.0 & 27.9 & 22.6 & 62.4 & \textbf{16.3} & 42.1 & 46.5 & \textbf{8.35} & \textbf{23.2} & \textbf{19.9} \\
Sup. Grd. +Cls. & 0 & 0.1 & 1 & 70.4 & 28.0 & \textbf{22.7} & 62.8 & \textbf{16.3} & 29.0 & \textbf{51.2} & 5.19 & 13.7 & 19.7 \\
Sup. Attn.+Grd.+Cls. & 0.1 & 0.1 & 1 & \textbf{70.6} & \textbf{28.1} & 22.6 & \textbf{63.3} & \textbf{16.3} & 41.2 & \textbf{50.8} & 8.30 & \textbf{23.2} & 19.6 \\
\bottomrule
\end{tabular}
}\vspace{-5pt}
\caption{Results on Flickr30k Entities val set. The top two scores on each metric are in bold.}\vspace{-10pt}
\label{tab:results_img}
\end{table*}

\begin{table}[t]
\centering
{\footnotesize
    \begin{tabular}{l|rr|rr}
    \toprule
    & \multicolumn{2}{c}{$\text{F1}_{all}$} & \multicolumn{2}{c}{$\text{F1}_{loc}$}\\
        Method &   Precision & Recall & Precision & Recall \\
        \midrule
        Unsup. (w/o SelfAttn) & 4.08 & 4.89 & 12.8 & 12.8  \\
        Unsup.  & 0.75 & 0.87 & 2.08 & 2.10  \\
        Sup. Attn.  & 7.46 & 8.83 & 22.4 & 22.5  \\
        Sup. Grd.   & 6.90 & 8.43 & 21.0 & 21.0  \\
        Sup. Cls. (w/o SelfAttn)  & 3.70 & 4.66 & 11.4 & 11.5  \\
        Sup. Attn.+Grd.  & 7.93 & 9.45 & 23.7 & 23.6  \\
        Sup. Attn.+Cls.  & 7.61 & 9.25 & 23.2 & 23.1  \\
        Sup. Grd. +Cls.  & 4.70 & 5.83 & 13.7 & 13.7  \\
        Sup. Attn.+Grd.+Cls.  & 7.56 & 9.20 & 23.2 & 23.2  \\
        \bottomrule
    \end{tabular}\vspace{-5pt}
    \caption{Attention precision and recall on generated sentences on Flickr30k Entities val set. All values are in \%.}
    \label{tab:flickr_pr_val}
}
\end{table}

\begin{table}[t]
\centering
{\footnotesize
    \begin{tabular}{l|rr|rr}
    \toprule
    & \multicolumn{2}{c}{$\text{F1}_{all}$} & \multicolumn{2}{c}{$\text{F1}_{loc}$}\\
        Method &   Precision & Recall & Precision & Recall \\
        \midrule
        BUTD~\cite{anderson2018bottom} & 4.07 & 5.13 & 13.1 & 13.0 \\
        Our Unsup. (w/o SelfAttn) & 3.44 & 4.47 & 11.6 & 11.8 \\
        Our Sup. Attn.+Grd.+Cls. & 6.91 & 8.33 & 22.2 & 22.2 \\
        \bottomrule
    \end{tabular}\vspace{-5pt}
    \caption{Attention precision and recall on generated sentences on Flickr30k Entities test set. All values are in \%.}\vspace{-10pt}
    \label{tab:flickr_pr_test}
}
\end{table}

\head{Qualitative examples.} See Figs.~\ref{fig:anet_vis1} and~\ref{fig:anet_vis2} for qualitative results of our methods and the Masked Transformer on \ourdatasetabbrev val set. We visualize the proposal with the highest attention weight in the corresponding frame. In (a), the supervised model correctly attends to ``man'' and ``Christmas tree'' in the video when generating the corresponding words. The unsupervised model mistakenly predicts ``Two boys''. In (b), both ``man'' and ``woman'' are correctly grounded. In (c), both ``man'' and ``saxophone'' are correctly grounded by our supervised model while Masked Transformer hallucinates a ``bed''. In (d), all the object words (\ie, ``people'', ``beach'', ``horses'') are correctly localized. The caption generated by Masked Transformer is incomplete. In (e), surprisingly, not only major objects ``woman'' and ``court'' are localized, but also the small object ``ball'' is attended with a high precision. Masked Transformer incorrectly predicts the gender of the person.
In (f), the Masked Transformer outputs an unnatural caption ``A group of people are in a raft and a man in red raft raft raft raft raft'' containing consecutive repeated words ``raft''.

\begin{figure*}[t]
\begin{minipage}[t]{1\textwidth}
\includegraphics[width=\textwidth]{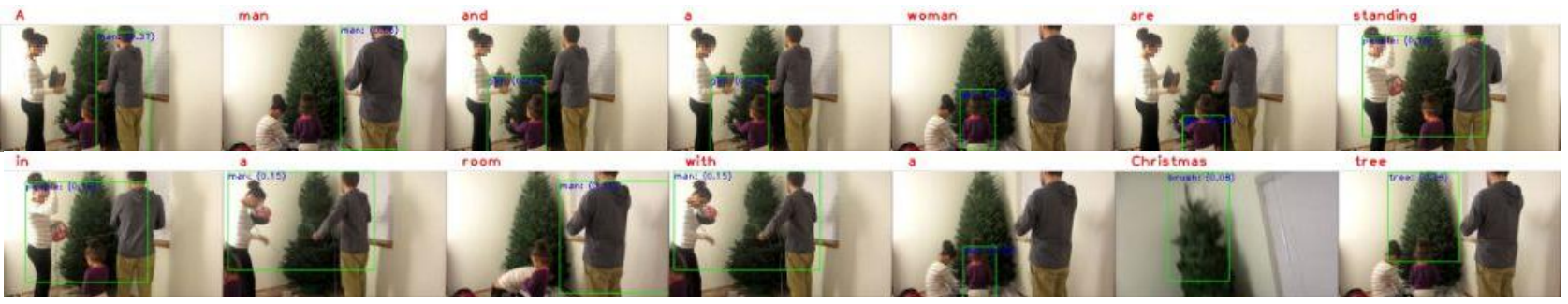} \\ (a) \textbf{Sup.}: A man and a woman are standing in a room with a Christmas tree; \\ \textbf{Unsup.}: Two boys are seen standing around a room holding a tree and speaking to one another; \\ \textbf{Masked Trans.}: They are standing in front of the christmas tree; \\ \textbf{GT}: Then, a man and a woman set up a Christmas tree.
\end{minipage}
\begin{minipage}[t]{1\textwidth}
\includegraphics[width=\textwidth]{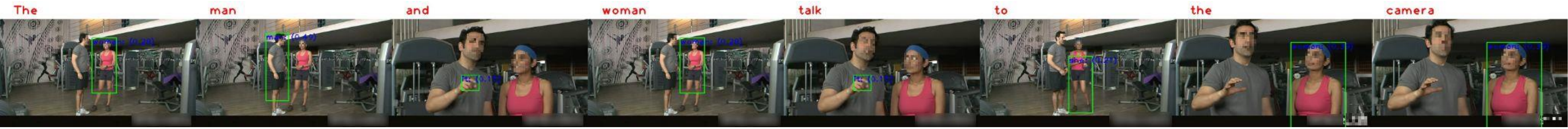} \\ (b) \textbf{Sup.}: The man and woman talk to the camera; \\ \textbf{Unsup.}: The man in the blue shirt is talking to the camera; \\ \textbf{Masked Trans.}: The man continues speaking while the woman speaks to the camera; \\ \textbf{GT}: The man and woman continue speaking to the camera.
\end{minipage}
\begin{minipage}[t]{1\textwidth}
\includegraphics[width=\textwidth]{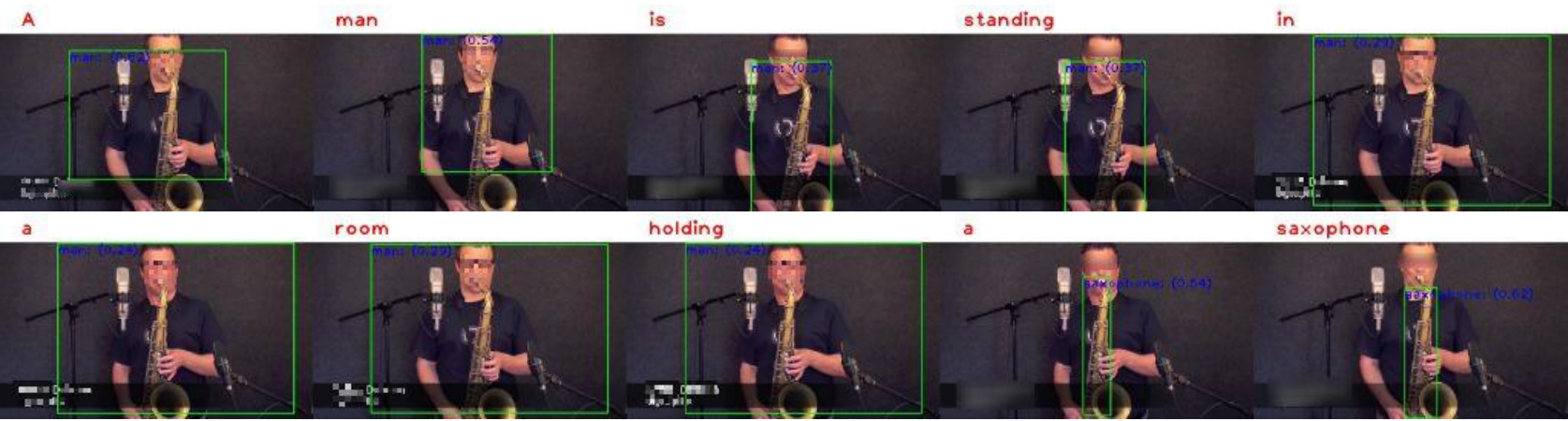} \\ (c) \textbf{Sup.}: A man is standing in a room holding a saxophone; \\ \textbf{Unsup.}: A man is playing a saxophone; \\ \textbf{Masked Trans.}: A man is seated on a bed; \\ \textbf{GT}: We see a man playing a saxophone in front of microphones.
\end{minipage}
\begin{minipage}[t]{1\textwidth}
\includegraphics[width=\textwidth]{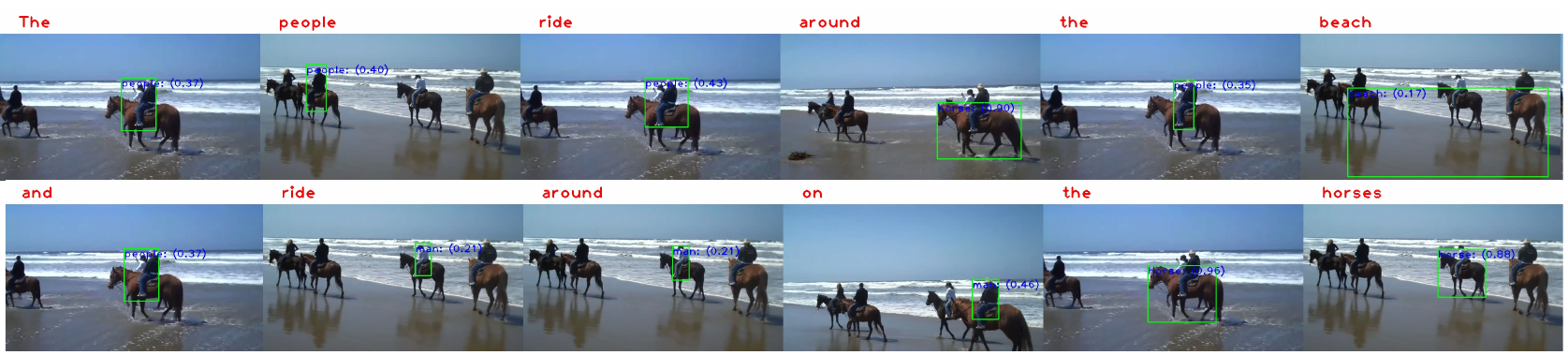} \\ (d) \textbf{Sup.}: The people ride around the beach and ride around on the horses; \\ \textbf{Unsup.}: The people ride around the beach and ride around; \\ \textbf{Masked Trans.}: The camera pans around the area and the girl leading the horse and the woman leading the; \\ \textbf{GT}: We see four people on horses on the beach.
\end{minipage}
\caption{Qualitative results on \ourdatasetabbrev val set. The red text at each frame indicates the generated word. The green box indicates the proposal with the highest attention weight. The blue text inside the green box corresponds to i) the object class with the highest probability and ii) the attention weight. Better zoomed and viewed in color. See Sec. \ref{sec:results:activitynet} for discussion.}
\label{fig:anet_vis1}
\end{figure*}

\begin{figure*}[t]
\begin{minipage}[t]{1\textwidth}
\includegraphics[width=\textwidth]{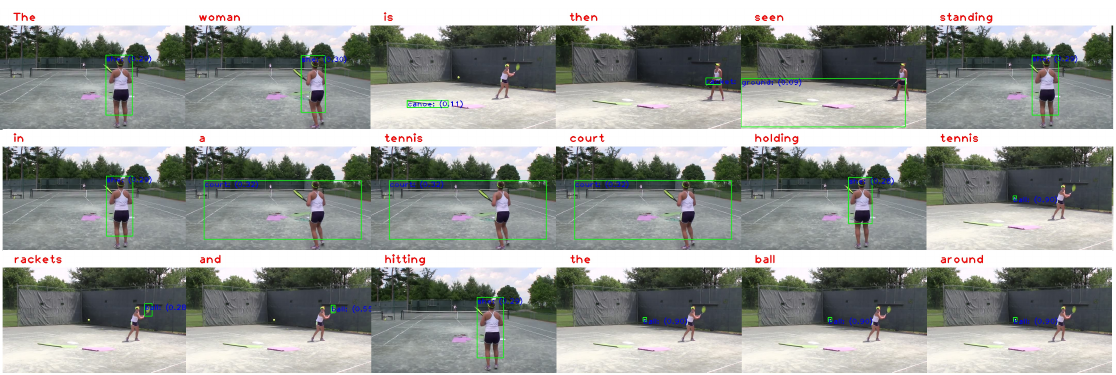} \\ (e) \textbf{Sup.}: The woman is then seen standing in a tennis court holding tennis rackets and hitting the ball around; \\ \textbf{Unsup.}: The woman serves the ball with a tennis racket; \\ \textbf{Masked Trans.}: We see a man playing tennis in a court; \\ \textbf{GT}: Two women are on a tennis court, showing the technique to posing and hitting the ball.
\end{minipage}
\begin{minipage}[t]{1\textwidth}
\includegraphics[width=\textwidth]{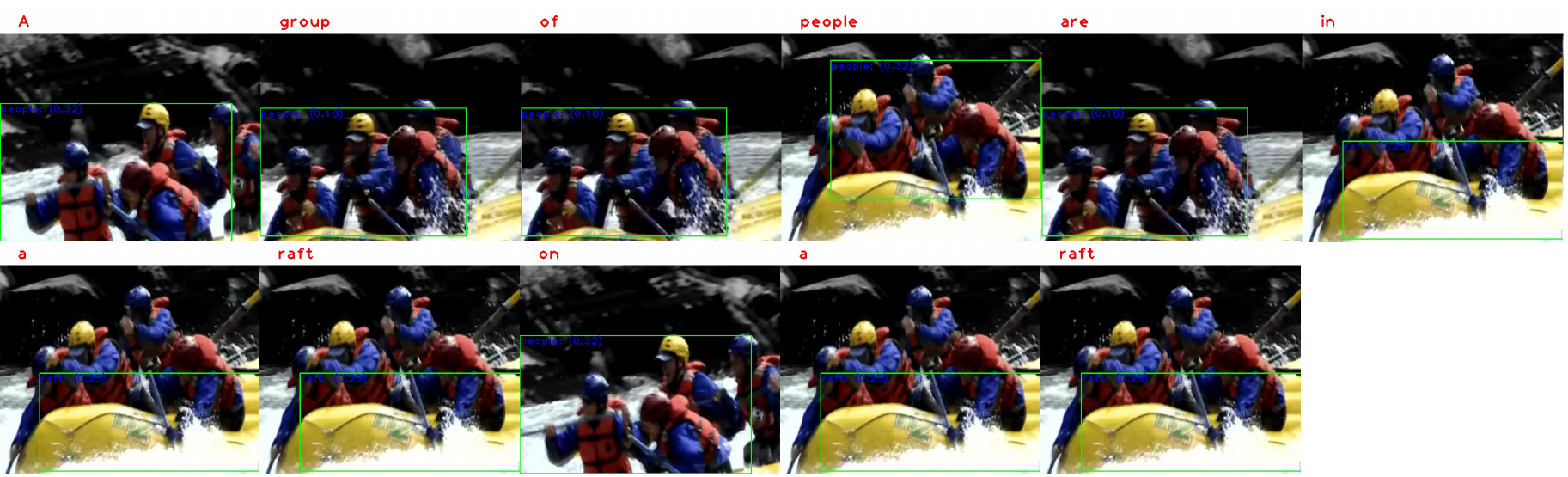} \\ (f) \textbf{Sup.}: A group of people are in a raft on a raft; \\ \textbf{Unsup.}: A group of people are in a raft; \\ \textbf{Masked Trans.}: A group of people are in a raft and a man in red raft raft raft raft raft; \\ \textbf{GT}: People are going down a river in a raft.
\end{minipage}
\caption{(Continued) Qualitative results on \ourdatasetabbrev val set. See the caption in Fig.~\ref{fig:anet_vis1} for more details.}
\label{fig:anet_vis2}
\end{figure*}

\subsection{Results on Flickr30k Entities}
\label{sec:results:flickr30k}
See Tab.~\ref{tab:results_img} for the results on Flickr30k Entities val set. Note that the results on the test set can be found in the main paper in Tab.~\ref{tab:results_img_test}. The proposal upper bound for attention and grounding is 90.0\%. For supervised methods, we perform a light hyper-parameter search and notice the setting $\lambda_{\alpha}=0.1$, $\lambda_{\beta}=0.1$ and $\lambda_{c}=1$ generally works well. The supervised methods outperform the unsupervised baseline by a decent amount in all the metrics with only one exceptions: Sup. Cls., which has a slightly inferior result in CIDEr. The best supervised method outperforms the best unsupervised baseline by a relative 0.9-4.8\% over all the metrics. The precision and recall associated with $\text{F1}_{all}$ and $\text{F1}_{loc}$ are shown in Tabs.~\ref{tab:flickr_pr_val}, \ref{tab:flickr_pr_test}.

\begin{figure*}[t]
\begin{minipage}[t]{1\textwidth}
\includegraphics[width=0.9\textwidth]{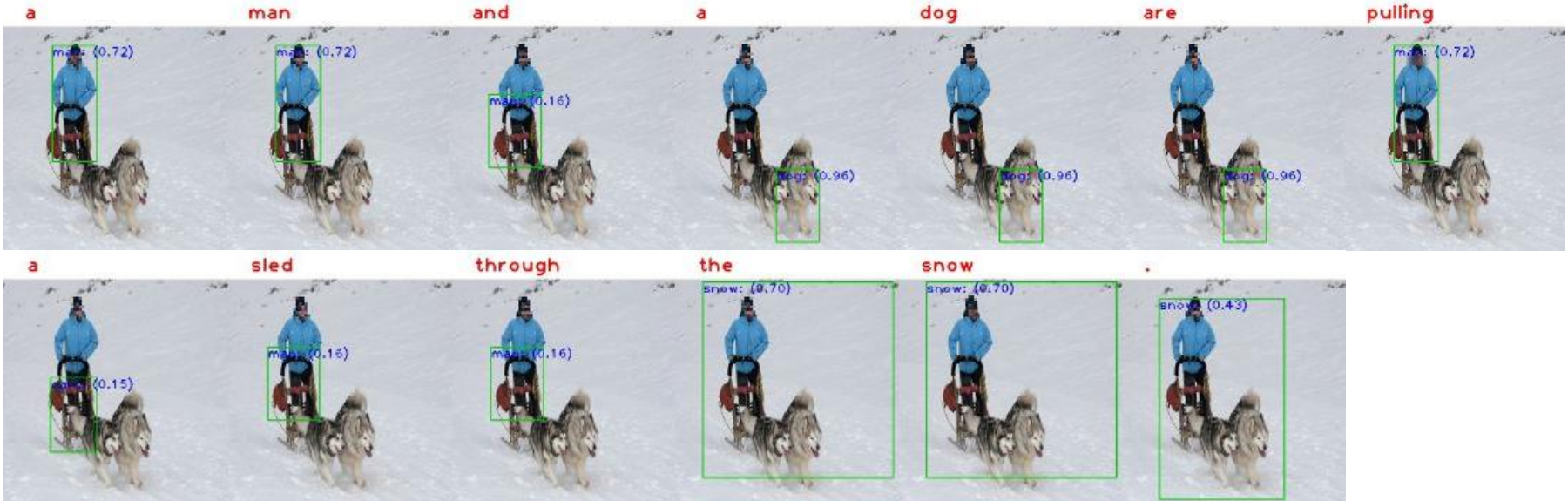} \\ (a) \textbf{Sup.}: A man and a dog are pulling a sled through the snow; \\ \textbf{Unsup.}: A man in a blue jacket is pulling a dog on a sled; \\ \textbf{BUTD}: Two dogs are playing in the snow; \\ \textbf{GT (5)}: A bearded man wearing a blue jacket rides his snow sled pulled by his two dogs / Man in blue coat is being pulled in a dog sled by two dogs / A man in a blue coat is propelled on his sled by two dogs / A man us using his two dogs to sled across the snow / Two Huskies pull a sled with a man in a blue jacket.
\end{minipage}
\begin{minipage}[t]{1\textwidth}
\includegraphics[width=0.6\textwidth]{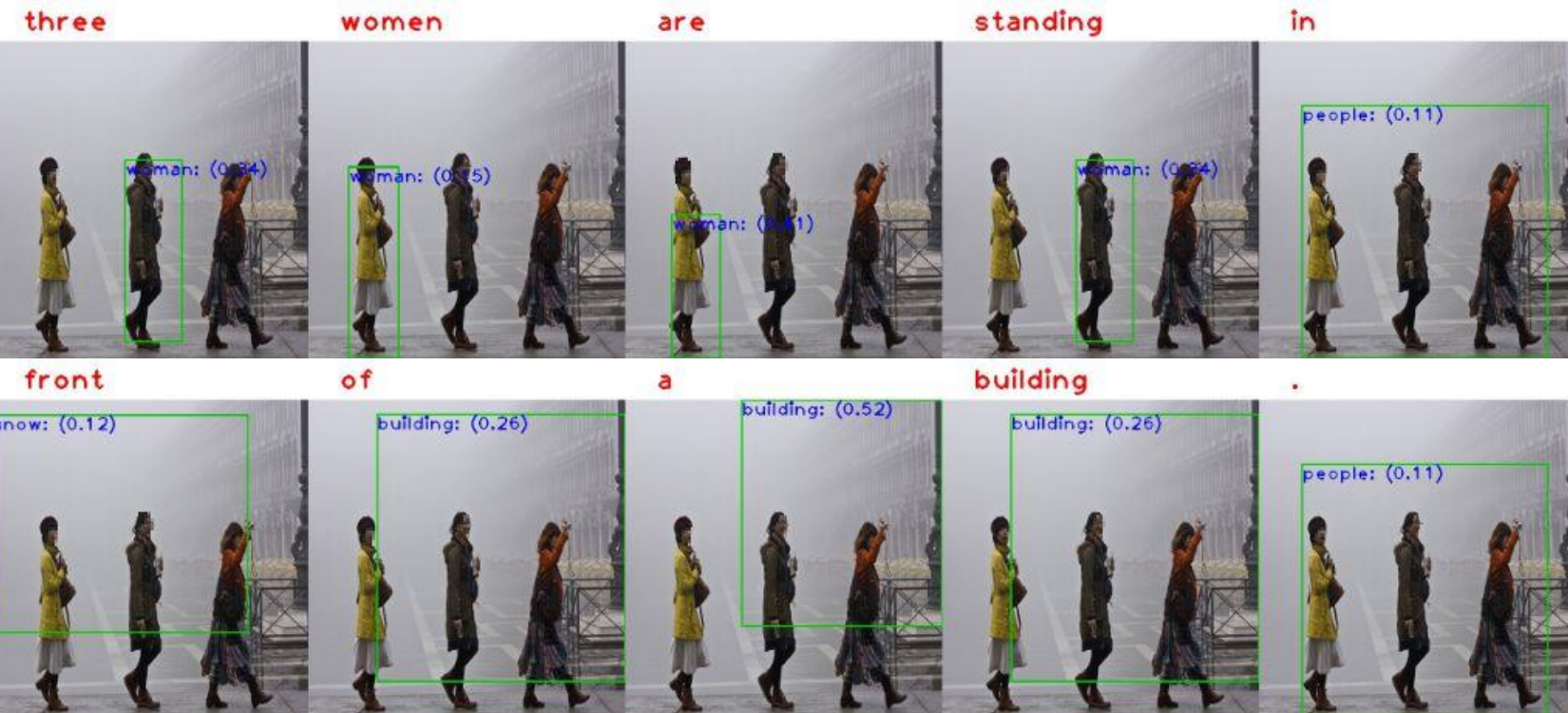} \\ (b) \textbf{Sup.}: Three women are standing in front of a building; \\ \textbf{Unsup.}: Three women in costumes are standing on a stage with a large wall in the background; \\ \textbf{BUTD}: Three women in yellow and white dresses are walking down a street; \\ \textbf{GT (5)}: Three woman are crossing the street and on is wearing a yellow coat / Three ladies enjoying a stroll on a cold, foggy day / A woman in a yellow jacket following two other women / Three women in jackets walk across the street / Three women are crossing a street.
\end{minipage}
\begin{minipage}[t]{1\textwidth}
\includegraphics[width=1\textwidth]{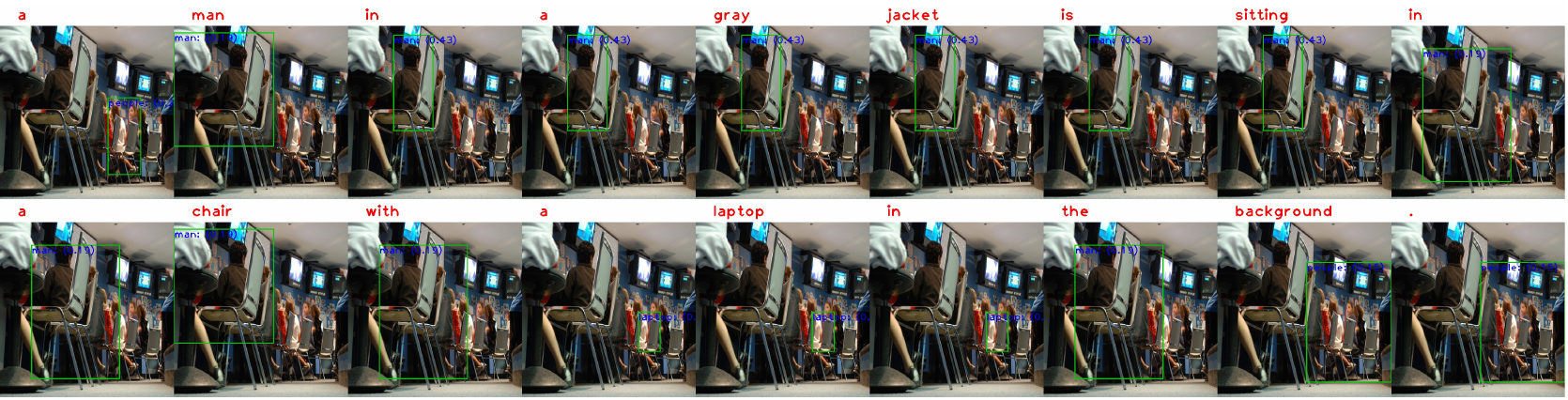} \\ (c) \textbf{Sup.}: A man in a gray jacket is sitting in a chair with a laptop in the background; \\ \textbf{Unsup.}: A man in a brown jacket is sitting in a chair at a table; \\ \textbf{BUTD}: A man in a brown jacket is sitting in a chair with a woman in a brown jacket in a; \\ \textbf{GT (5)}: Several chairs lined against a wall, with children sitting in them / A group of children sitting in chairs with monitors over them / Children are sitting in chairs under some television sets / Pre-teen students attend a computer class / Kids conversing and learning in class.
\end{minipage}
\caption{Qualitative results on Flickr30k Entities val set. Better zoomed and viewed in color. See Sec. \ref{sec:results:flickr30k} for discussion.}
\label{fig:flickr_vis}
\end{figure*}

\head{Qualitative examples.} See Fig.~\ref{fig:flickr_vis} for the qualitative results by our methods and the BUTD on Flickr30k Entities val set. We visualize the proposal with the highest attention weight as the green box. The corresponding attention weight and the most confident object prediction of the proposal are displayed as the blue text inside the green box. In (a), the supervised model correctly attends to ``man'', ``dog'' and ``snow'' in the image when generating the corresponding words. The unsupervised model misses the word ``snow'' and BUTD misses the word ``man''. In (b), the supervised model successfully incorporates the detected visual clues (\ie, ``women'', ``building'') into the description. We also show a negative example in (c), where interestingly, the back of the chair looks like a laptop, which confuses our grounding module. The supervised model hallucinates a ``laptop'' in the scene.

\subsection{Implementation Details}
\label{sec:implementation:details}
\head{Region proposal and feature.} We uniformly sample 10 frames per video segment (an event in \ourdatasetabbrev) and extract region features. For each frame, we use a Faster RCNN model~\cite{ren2015faster} with a ResNeXt-101 FPN backbone~\cite{xie2017aggregated} for region proposal and feature extraction. The Faster RCNN model is pretrained on the Visual Genonme dataset~\cite{krishna2017visual}. We use the same train-val-test split pre-processed by Anderson \etal~\cite{anderson2018bottom} for joint object detection (1600 classes) and attribute classification.
In order for a proposal to be considered valid, its confident score has to be greater than 0.2. And we limit the number of regions per image to a fixed 100~\cite{jiang2018pythia}. We take the output of the fc6 layer as the feature representation for each region, and fine-tune the fc7 layer and object classifiers with 0.1$\times$ learning rate during model training. \\

\head{Training details.} We optimize the training with Adam (params: 0.9, 0.999). The learning rate is set to 5e-4 in general and to 5e-5 for fine-tuning, \ie, fc7 layer and object classifiers, decayed by 0.8 every 3 epochs. The batch size is 240 for all the methods. We implement the model in PyTorch based on NBT\footnote{https://github.com/jiasenlu/NeuralBabyTalk} and train on 8x V100 GPUs. The training is limited to 40 epochs and the model with the best validation CIDEr score is selected for testing. \\

\begin{table*}[t]
\begin{tabular}{|l|l|l|l|l|l|l|l|}
\toprule
\_\_background\_\_ & egg & nail & kid & snowboard & hoop & roller & pasta \\
bagpipe & stilt & metal & butter & cheerleader & puck & kitchen & stage \\
coach & paper & dog & surfboard & landscape & scene & guitar & trophy \\
bull & dough & tooth & object & eye & scissors & grass & stone \\
rod & costume & pipe & ocean & sweater & ring & drum & swimmer \\
disc & oven & shop & person & camera & city & accordion & stand \\
dish & braid & shot & edge & vehicle & horse & ramp & road \\
chair & pinata & kite & bottle & raft & basketball & bridge & swimming \\
carpet & bunch & text & camel & themselves & monkey & wall & image \\
animal & group & barbell & photo & calf & top & soap & playground \\
gymnast & harmonica & biker & polish & teen & paint & pot & brush \\
mower & platform & shoe & cup & door & leash & pole & female \\
bike & window & ground & sky & plant & store & dancer & log \\
curler & soccer & tire & lake & glass & beard & table & area \\
ingredient & coffee & title & bench & flag & gear & boat & tennis \\
woman & someone & winner & color & adult & shorts & bathroom & lot \\
string & sword & bush & pile & baby & gym & teammate & suit \\
wave & food & wood & location & hole & wax & instrument & opponent \\
gun & material & tape & ski & circle & park & blower & head \\
item & number & hockey & skier & word & part & beer & himself \\
sand & band & piano & couple & room & herself & stadium & t-shirt \\
saxophone & they & goalie & dart & car & chef & board & cloth \\
team & foot & pumpkin & sumo & athlete & target & website & line \\
sidewalk & silver & hip & game & blade & instruction & arena & ear \\
razor & bread & plate & dryer & roof & tree & referee & he \\
clothes & name & cube & background & cat & bed & fire & hair \\
bicycle & slide & beam & vacuum & wrestler & friend & worker & slope \\
fence & arrow & hedge & judge & closing & iron & child & potato \\
sign & rock & bat & lady & male & coat & bmx & bucket \\
jump & side & bar & furniture & dress & scuba & instructor & cake \\
street & everyone & artist & shoulder & court & rag & tank & piece \\
video & weight & bag & towel & goal & clip & hat & pin \\
paddle & series & she & gift & clothing & runner & rope & intro \\
uniform & fish & river & javelin & machine & mountain & balance & home \\
supplies & gymnasium & view & glove & rubik & microphone & canoe & ax \\
net & logo & set & rider & tile & angle & it & face \\
exercise & girl & frame & audience & toddler & snow & surface & pit \\
body & living & individual & crowd & beach & couch & player & cream \\
trampoline & flower & parking & people & product & equipment & cone & lemon \\
leg & container & racket & back & sandwich & chest & violin & floor \\
surfer & house & close & sponge & mat & contact & helmet & fencing \\
water & hill & arm & mirror & tattoo & lip & shirt & field \\
studio & wallpaper & reporter & diving & ladder & tool & paw & other \\
sink & dirt & its & slice & bumper & spectator & bowl & oar \\
path & toy & score & leaf & end & track & member & picture \\
box & cookie & finger & bottom & baton & flute & belly & frisbee \\
boy & guy & teens & tube & man & cigarette & vegetable & lens \\
stair & card & pants & ice & tomato & mouth & pan & pool \\
bow & yard & opening & skateboarder & neck & letter & wheel & building \\
credit & skateboard & screen & christmas & liquid & darts & ball & lane \\
smoke & thing & outfit & knife & light & pair & drink & phone \\
trainer & swing & toothbrush & hose & counter & knee & hand & mask \\
shovel & castle & news & bowling & volleyball & class & fruit & jacket \\
kayak & cheese & tub & diver & truck & lawn & student & stick \\
\bottomrule
\end{tabular}\label{tab:list_of_obj}
\caption{List of objects in \ourdataset, including the ``\_\_background\_\_'' class.}
\end{table*}

\end{document}